**Integration of the DOLCE top-level ontology
into the OntoSpec methodology**

**Gilles KASSEL[a]**

a LaRIA, University of Picardie Jules Verne, gilles.kassel@u-picardie.fr

# Integration of the DOLCE top-level ontology
# into the OntoSpec methodology [*]

Gilles Kassel

LaRIA, Knowledge Engineering Team
University of Picardie Jules Verne
33 rue Saint-Leu
F-80039 Amiens cedex 1
France
gilles.kassel@u-picardie.fr

## Abstract

This report describes a new version of the OntoSpec methodology for ontology building. Defined by the LaRIA Knowledge Engineering Team (University of Picardie Jules Verne, Amiens, France), OntoSpec aims at helping builders to model ontological knowledge (upstream of formal representation). The methodology relies on a set of rigorously-defined modelling primitives and principles. Its application leads to the elaboration of a semi-informal ontology, which is independent of knowledge representation languages. We recently enriched the OntoSpec methodology by endowing it with a new resource, the DOLCE top-level ontology defined at the LOA (IST-CNR, Trento, Italy). The goal of this integration is to provide modellers with additional help in structuring application ontologies, while maintaining independence vis-à-vis formal representation languages. In this report, we first provide an overview of the OntoSpec methodology's general principles and then describe the DOLCE re-engineering process. A complete version of DOLCE-OS (i.e. a specification of DOLCE in the semi-informal OntoSpec language) is presented in an appendix.

# 1 Introduction

The work presented in this report aims at completing the definition of the OntoSpec methodology for ontology building [Kassel, 2002] by endowing it with an ontological resource - the DOLCE top-level ontology [Masolo *et al.*, 2003].

## 1.1 The OntoSpec methodology

The OntoSpec methodology provides the ontology builder with a modelling framework which allows him/her to define (via successive refinements) the conceptual entities (concepts and relations) composing the ontology, by first identifying and then progressively modelling the properties characterising these entities.

In line with logical tradition, the framework mainly consists of a set of "roles" corresponding to the different ways in which properties contribute to the definition of other properties (e.g., necessary membership condition, necessary and sufficient membership condition). This framework integrates Nicola Guarino and Christopher Welty's recent

---

suggestions of using notions originating from the discipline of Formal Ontology (e.g. identity condition, unity condition) in order to complete the definition of properties [Guarino and Welty, 2000a, 2000b][Welty and Guarino, 2001].

The general principle of OntoSpec (as a methodology) consists in identifying ever more precise roles defined in a generalisation/specialisation taxonomy, whilst taking into account the structure of the properties in question. Therefore, a concept property which is initially qualified as a "necessary membership condition" can thereafter be characterised by the builder as a "subsumption link with differentia" or an "existential restriction". In the same way, a "necessary membership condition" defining a relation can subsequently be characterised as a simple "subsumption link" or as a "domain or range restriction".

OntoSpec's contribution comes through helping the builder with the ontological knowledge modelling step, upstream of the formal representation and knowledge implementation steps. Hence, along with the METHONTOLOGY [Fernández-López *et al.*, 1999], TERMINAE [Aussenac-Gilles *et al.*, 2000] and ARCHONTE [Bachimont, 2004] methodologies, OntoSpec's concern is first to model a conceptualisation at the "knowledge level". OntoSpec thus stands apart from methodologies (such as that attached to the PROTÉGÉ environment [Noy and McGuinness, 2001]), which propose primitives from object-oriented programming languages (object, slot, value) as their modelling primitives[1].

OntoSpec uses controlled and highly structured natural language as a specification mode. The structure corresponds to the above-mentioned roles and it is specified by way of labels attached to the natural language statements of properties. An ontology defined in OntoSpec is therefore a "semi-informal" ontology, according to the typology of [Ushold and Grüninger, 1996]. OntoSpec is thus independent of formal representation languages. Consequently, its definitions remain understandable by all, which enables experts in the given domain or future users of the ontology to cooperate with the builder by evaluating the modelling choices and the quality of the resulting definitions.

The OntoSpec method has recently been integrated (from a conceptual point of view) into the TERMINAE methodology for text-based ontology building [Aussenac-Gilles *et al.*, 2000], as part of the French ATONANT RNTL-Technolangue project [Ben Khédija, 2004]. Its integration (in terms of software) into the TERMINAE platform is currently underway [Bruaux, 2005]. In this report, we describe additional work aimed at endowing the methodology with a top level ontology in order to help build application ontologies.

## 1.2 The DOLCE ontology

For this project, we needed a top level ontology that would constitute a good starting point for elaborating new ontologies. The candidates included the SUMO[2] ontology (Suggested Upper Merged Ontology), produced by the IEEE SUO project [Niles and Pease, 2001], and the DOLCE ontology (Descriptive Ontology for Linguistic and Cognitive Engineering), one of the outputs of the IST WonderWeb 2001-2003 project[3], in which various members of the LOA (Laboratory for Applied Ontology, ISTC-CNR, Trento, Italy)[4] participated. We finally chose the DOLCE ontology, for the following reasons:

---

[1] More generally, while introducing a level of "conceptual" modelling (or a "knowledge level"), OntoSpec stands out from ontology construction methods whereby editors displaying formal representations of ontologies are used directly - practices advocated, for example, by the authors of the WebOnto [Domingue, 1998] and OilEd [Bechhofer *et al.*, 2001] editors.
[2] http://ontology.teknowledge.com/.
[3] http://wonderweb.semanticweb.org/.
[4] http://www.loa-cnr.it/.

- DOLCE possesses a rich axiom set, which allows one to access the "ontological commitments" on which the ontology is based. Its reuse is thus facilitated.
- The modelling choices for structuring the ontology are explicit and justified, which also facilitates reuse.
- Thanks to its direct relationship with the works of Nicola Guarino and Christopher Welty cited in section 1.1, DOLCE comprises a set of abstract, "rigids"[5] concepts, which are considered to be both necessary and sufficient for subsuming any concept in any application ontology. Its extension is therefore facilitated.
- The DOLCE ontology is the subject of ongoing work and a number of extensions have been provided, in particular in the form of core ontologies[6] which correspond to generic domain ontologies. In the near future, we expect to integrate these extensions into OntoSpec.

Several versions of the DOLCE ontology (specified in different languages) exist. Indeed, DOLCE was originally specified in first-order modal logic and then represented in languages such as KIF3.0, DAML+OIL and OWL. In the present work, we have used the original version (specified as an axiomatic theory in modal logic) as our reference. The latter was thoroughly presented in the reference [Masolo *et al.*, 2003] - which we will name D18 in the rest of the report. The original version is the richest (from a semantic point of view), because it is represented in the language offering the greatest expressive power. This choice is indeed consistent with the use of DOLCE in a conceptual modelling step, where it is above all a matter of conferring meaning. In this report, we shall also see that translation into the semi-informal OntoSpec language was carried out without losing expressive power.

**1.3 Structure of the report**

In the rest of this report, we first present an overview of the OntoSpec methodology (section 2), then describe the DOLCE re-engineering process (section 3), according to the steps recommended by OntoSpec. In an appendix, we first give the list of conditions constituting the OntoSpec modelling framework and then present a complete version of DOLCE-OS, a specification of DOLCE in the semi-informal OntoSpec language.

## 2 Overview of OntoSpec

In terms of input data, the OntoSpec method accepts a set of conceptual entities expressed by terms, together with a set of natural language texts defining the said entities. The data may originate from the linguistic analysis of documents, as is the case in the TERMINAE-OntoSpec pairing. As a result, OntoSpec allows the elaboration of semi-informal definitions which are amenable to being coded in a formal representation language.

The process for transformation of natural language texts into a semi-informal ontology is composed of five main steps, all of which will be reviewed in this section. These steps apply to the definition of each conceptual entity: the distinction in the text between a "definition" (corresponding to the statement of properties being satisfied by the instances of the conceptual entity) on one hand and a "comment" on the other (2.1); the distinction between *essential* properties and *contingent* properties (2.2); for each property, the identification of its role vis-à-vis the defined entity, in terms of *conditions* (2.3); the explicit formulation of the structure of the membership conditions, in order to identify more precise conditions (2.4) and the attribution of meta-properties to the defined conceptual entity, according to the OntoClean's principles [Guarino and Welty, 2004] (2.5). In parallel with these steps, the comment part is also structured by recognizing recurrent categories of comments (2.6).

---

[5] This term is defined later in the report.
[6] These ontologies were the subject of the CORONT (CORe ONTologies) Workshop at the EKAW'2004 conference (http://www.loa-cnr.it/coreont.html), the proceedings of which can be accessed at the following address: http://CEUR-WS.org/vol-118.

For each step, we first state the notions brought into play and then illustrate its implementation using a series of examples. Rather than suggesting a particular conceptualisation of the world, the examples seek to illustrate the step's contribution to the overall structuring of the ontology. As one progresses from one step to the next, the definitions become increasingly structured.

**2.1 Definition of a conceptual entity**

*2.1.1 Notions*

Since an ontology is generally defined *a minima* as a "specification of a conceptualisation", the modelling framework of OntoSpec first comprises a set of primitives (e.g. property, concept, relation) which enable elaboration of a conceptualisation:
- A *property* is the *meaning*, or *intension*, of a linguistic expression such as "being a car" or "being the tenant of the ground floor apartment n°3", or "eating". One important characteristic of *properties* is that they *classify* entities belonging to a world, or *instances*. The set of instances classified by a property (i.e. those which verify or satisfy the property) constitutes the *reference* or *realisation*, or indeed *extension*, of the property.
- Within properties, one can distinguish *concepts* (which have a set of individual instances or *individuals* as their extension) and *relations* (which have a set of tuples of individuals as their extension). Formally, in first order logic, a concept is represented by a unary predicate whereas a relation is represented by a n-ary (n ≥ 2) predicate.
- Again within properties, one can distinguish *meta-properties*, which have a set of properties as their extension. Some of these meta-properties are concepts and others are relations.

We have nevertheless added a constraint concerning the notion of concept: two co-referential concepts, with the same extension all the time, must be identical. This constraint aims at simplifying the structure of the system of concepts constituting a conceptualisation in OntoSpec. Therefore, the meaning of expressions such as "a geometric figure with three sides" and "a geometric figure with three angles" are considered as constituting only one concept - the intension of the class of figures named "triangles". This constraint entails that our notion of concept is consistent with the notion of *universal* in DOLCE and also with the notion of *universal* formalised by [Bittner *et al.*, 2004].

These definitions allow us to specify our acceptation of the term "ontology". According to OntoSpec, an ontology is defined as a set of conceptual entities, each characterized by one (or several) term(s) and a definition. The definition of a conceptual entity corresponds to natural language specification of its intension. It consists of a text in three main parts:
- A statement of properties satisfied by all the instances classified by the conceptual entity (the entity's extension).
- A statement of meta-properties satisfied by the conceptual entity.
- A comment on the first two parts which aims at facilitating their understanding and adoption.

Strictly speaking, only the first two parts of the definition are really "definitional" by helping characterize the intension of the conceptual entity. Henceforth, the term "definition" will refer only to these parts. The distinction made between the comment part and its separate expression constitutes the first step of the OntoSpec method.

*2.1.2 Examples*

Here we illustrate this distinction by way of examples, bearing in mind that at this stage, the definition does not comprise statements of meta-properties. These examples correspond to informal definitions extracted from the SUMO ontology [Pease and Niles, 2002] and within which one can recognize the two parts.

Example 1:

**Concept**: Region
  **Definition**
    A region is a topographical location.
  **Comment**
    Regions encompass surfaces of objects, imaginary places and geographic areas. Note that a region is the only kind of object which can be located at itself. Note too that "region" is not a subclass of "self-connected object", because some regions (e.g. archipelagos) have parts which are not connected with one another.

Example 2:

**Concept**: Battle
  **Definition**
    A battle is a violent contest between two or more military units within the context of a war.
  **Comment**
    Note that this does not cover the metaphorical sense of "battle", which simply means a struggle of some sort. This sense should be represented with the more general concept of "contest".

In the SUMO ontology, these informal definitions are accompanied by formal definitions (specified in the suo-KIF language). The contribution of OntoSpec is precisely that of helping the builder in his/her knowledge modelling work, in order to obtain this type of formal definition.

## 2.2 Distinction of *essential vs contingent* properties

*2.2.1 Notions*

In line with a convention dating back to Aristotle, two types of predication (or attribution of a predicate to a subject) have been distinguished: one type in which the subject and the predicate are in an *essential* relation (e.g. "Socrates is a human") and another in which these two entities are in an *incidental* or *accidental* relation (e.g. "Socrates is sitting"). Today, this distinction is at the heart of ontology construction:

> [Bouaud *et al.*, 1994]: "A careful distinction must be made between *essential* and *incidental*, i.e. non essential, properties. This distinction is important to clarify what must be considered as the basic meaning of a type: its essence. The essential characteristics of objects are those such that, if a defined object looses only one of them, it no longer exists as such. These properties must be true by intension as long as the object exists. These properties are definitional, in the sense that objects that carry them are recognized as members of the type in every possible world."

As this citation shows, it is customary to account for this distinction by appealing to the theory of *possible worlds*, which constitutes the foundation of the semantics in propositional modal logics[7]. These logics distinguish between *necessary* propositions (true in every possible world) and *possible* propositions (true in at least one world).

The propositions of interest in the definitions here, are equivalent to properties attributed to the instances of the conceptual entity being defined. According to the modality of the proposition (i.e. necessary or possible), we consider two classes of properties: *essential* properties and *contingent* properties.

Definition 1: A property $\psi$ is *essential* for $\varphi$ if and only if the proposition attributing the property $\psi$ to all the instances of $\varphi$ is necessary.

Definition 2: A property $\psi$ is *contingent* for $\varphi$ if and only if the proposition attributing the property $\psi$ to all the instances of $\varphi$ is possible, but not necessary.

Comment: It is important to note that, according to these definitions, a property is not intrinsically essential or contingent[8]: it gains this status from its relation to another property.

According to certain authors, an ontology is only concerned with necessary conditions, that is to say solely with identification of essential properties for conceptual entities:

> [Guarino and Giaretta, 1995]: "An ontological theory contains formulas which are considered to be always true (and therefore sharable among multiple agents), independently of particular states of affairs. Formally, we can say that such formulas must be true in every possible world."

While not refuting this point of view in OntoSpec, we decided to include contingent properties in the definition of a conceptual entity - taking care, of course, to distinguish them from essential properties. The main goal is to force the modeller to pose the question of the properties' status, and thus to identify the truly defining properties. Moreover, while authorizing the specification of contingent properties, OntoSpec allows one to build knowledge bases which are broader than simple ontologies (but which often continue to be qualified as ontologies by their authors). The utility of this type of practice is emphasized in the following examples.

*2.2.2 Examples*

As an illustration, let us consider the example of the following definition of the term "confidential document"[9]: *A confidential document is a document which has been subjected to a classification procedure because it contains confidential information. Some of these*

---

[7] This semantic is usually attributed to the logician Saul Kripke. Accordingly, a possible world is a set of individuals and properties.
[8] In section 2.5, we define such meta-properties (notably by including the notion of *rigidity*) and relate them to the present notions.
[9] This definition is extracted from a corporate glossary.

*documents are classified as "confidential - corporate", whereas others are classified as "confidential - defence"*.

The first sentence is intended to be definitional and the second is a comment providing examples of documents considered as confidential. Although the first sentence follows the equation of the Aristotelian definition (species = genus + differentia) it does suggest that we may have overlooked the true definition: we are informed of the existence of a procedure, within the corporate body, that allows one to attribute "confidential" status to a document but, we still not really know what a *confidential* document is!

In essence, if we focus on properties satisfied by a confidential document in any possible setting, we are led to consider that it is a document which is divulged to a restricted readership, under the seal of secrecy. Furthermore, one perceives that, within this corporate body, the notions of "confidential corporate document" and "confidential defence document" are linked to restrictions on divulgation to different readerships. In contrast, the fact that *every document is subjected to a classification procedure* is contingent to this corporate body (corresponding to a sub-class of situations).

In terms of an application (a knowledge management system, for example), developed for this corporate body, it may be useful to memorise this information in a knowledge base. In the same way, in order to promote contacts between company employees, it may be interesting to memorise the fact that *every employee possesses a phone number and an electronic address*, even if these properties do not correspond to the essence of the concept of "employee".

Example 3:

**Concept**: Confidential document
  **Definition**
    **Essential properties**
      A confidential document is a document whose divulgation outside a defined readership is forbidden.
    **Contingent properties**
      Every confidential document is subjected to a classification procedure.
    **Comment**
      Some of these documents are classified as "confidential - corporate" and others are classified as "confidential - defence".

## 2.3 Distinction of categories of *conditions*

### 2.3.1 Notions

The essential and/or contingent properties forming part of the definition of another property help define, or characterise, the latter. We use the verb "to carry" to refer to this relationship of notional characterisation, which leads one to state that a property "carries" other properties or, conversely, that properties are "carried by" another property[10].

In line with a logical convention dating back to Carnap and Morris, it is customary to distinguish between different relationships of this type which, for the carried property, correspond to different ways of constraining the reference of the defined property. The stock term "condition", that we find in expressions such as "necessary condition" or "sufficient condition", conveys this notion of constraint.

---

[10] Here we adopt the terminology employed by Nicola Guarino and Christopher Welty in their articles (cf. [Welty and Guarino, 2001], Definition 5, for example).

The conditions used most frequently, because they correspond to the simplest logical forms, are *membership* conditions.

<u>Definition 3</u>: The property φ *carries* the property ψ as a *Necessary Membership Condition* (NMC) if and only if being an instance of φ implies being an instance of ψ.

The equivalent in logic of the statement "φ carries ψ as an NMC" is the following formula:

$\forall x \; \varphi(x) \rightarrow \psi(x)$; if φ and ψ are concepts.
$\forall x_1, x_2 \ldots, x_n \; \varphi(x_1, x_2 \ldots, x_n) \rightarrow \psi(x_1, x_2 \ldots, x_n)$; if φ and ψ are n-ary relations (n ≥ 2).

One must note that, according to our distinction between essential and contingent properties, we consider that a property φ can *essentially* or *contingently* carry a property ψ as an NMC[11]. This consideration also applies to the other categories of conditions defined in this section.

<u>Definition 4</u>: The property φ *carries* the property ψ as a *Sufficient Membership Condition* (SMC) if and only if being an instance of ψ implies being an instance of φ.

The equivalent in logic of the statement "φ carries ψ as an SMC" is the following formula:

$\forall x \; \psi(x) \rightarrow \varphi(x)$; if φ and ψ are concepts.
$\forall x_1, x_2 \ldots, x_n \; \psi(x_1, x_2 \ldots, x_n) \rightarrow \varphi(x_1, x_2 \ldots, x_n)$; if φ et ψ are n-ary relations (n ≥ 2).

<u>Definition 5</u>: The property φ *carries* the property ψ as a *Necessary and Sufficient Membership Condition* (NSMC) if and only if being an instance of φ is equivalent to being an instance of ψ.

The equivalent in logic of the statement "φ carries ψ as an NSMC" is the following formula:

$\forall x \; \psi(x) \leftrightarrow \varphi(x)$; if φ and ψ are concepts.
$\forall x_1, x_2 \ldots, x_n \; \psi(x_1, x_2 \ldots, x_n) \leftrightarrow \varphi(x_1, x_2 \ldots, x_n)$; if φ et ψ are n-ary relations (n ≥ 2).

In addition to membership conditions, OntoSpec considers *identity* and *unity* conditions, the importance of which for the definition of properties (and in particular for the analysis of subsumption links) has been emphasised in the works of Nicola Guarino and Christopher Welty [Guarino and Welty, 2000a, 2000b][Welty and Guarino, 2001]. Below, we cite the definitions given by these authors[12]. The reader should note that, in contrast to membership conditions (which can be carried by properties in general), identity and unity conditions only concern concepts.

<u>Definition 6</u>: The concept φ carries the relation ψ as a *Necessary and Sufficient Identity Condition* (NSIC) if and only if the relation ψ allows one to decide whether two instances of φ are identical.

---

[11] If Φ corresponds to the proposition *φ carries ψ as an NMC*, this is equivalent to considering that *Φ* is *necessary* or that *Φ* is *possible, but not necessary*.
[12] For comments on the origin of these notions but also on known difficulties concerning their usage, we refer the reader to the articles cited in the text.

The equivalent in logic of the statement "φ carries ψ as an NSIC" is the following formula:

$$\forall x,y \ (\varphi(x) \land \varphi(y) \rightarrow (\psi(x,y) \leftrightarrow x=y))$$

In some cases, the identity criteria ψ simply allows one to imply the instances' identity or, in contrast, is implied by the instances' identity: in such cases, ψ plays the role of a *Sufficient Identity Condition* (SIC) or a *Necessary Identity condition* (NIC).

Definition 7: The concept φ *carries* the relation ψ as a *Unity Condition* (UC) if and only if ψ is an equivalence relation such that each instance of φ constitutes a *whole* according to ψ, in the sense that all the parts of the instance – and only these parts – are related by ψ.

In the equivalent logical formula, we distinguish between *endurants*, *perdurants* and *abstracts* (in its meaning within DOLCE) because – in this ontology – these entities correspond to different parthood relations: temporal realations for the endurants and atemporal relations for the perdurants and abstracts. The expression ED(x) holds for "x *is an endurant*", PD(x) holds for "x *is a perdurant*", AB(x) holds for "x *is an abstract*", P(x,y) holds for "x *is a part of* y" and P(x,y,t) holds for "x *is a part of* y *at time* t".

$$\forall x,t \ \varphi(x) \rightarrow [(ED(x) \rightarrow (\forall y,z(P(y,x,t) \land P(z,x,t)) \rightarrow \psi(y,z)$$
$$\land \forall y,z(\neg P(y,x,t) \land \neg P(z,x,t)) \rightarrow \neg\psi(y,z)))$$
$$\land \ ((PD(x) \lor (AB(x)) \rightarrow (\forall y,z(P(y,x) \land P(z,x)) \rightarrow \psi(y,z)$$
$$\land \forall y,z(\neg P(y,x) \land \neg P(z,x)) \rightarrow \neg\psi(y,z)))]$$

*2.3.2 Examples*

The distinction between these different kinds of conditions enables us to label each property (be it essential or contingent) carried by the defined entity. We have also taken advantage of the examples below by specifying the OntoSpec's current notation. In particular, the labels appear between square brackets and comprise first an indication of the type of property ("EP", for essential property; "CP", for contingent property) and then an indication of the type of condition (cf. Example 4).

Example 4 (following on from Example 3):

**Concept**: Confidential document
  **Definition**
    **[EP/NSMC]** A confidential document is a document whose divulgation outside a fixed readership is forbidden.
    **[CP/NMC]** All confidential documents are subjected to a classification procedure.

Examples 5 and 6 illustrate the statement of an identity condition and a unity condition, respectively. According to the stated conditions, one can deduce (Example 5) that a topographical location not having exactly the same parts cannot correspond to the same region, and (Example 6) that the group of organs on which the notion of respiratory system is founded has a purely functional justification (rather than a topological justification, as in the previous example), since the organs contribute to the same physiological function. These two examples illustrate the fact that the statement of identity and unity conditions does indeed contributes to the definition of the present notions.

Example 5 (following on from Example 1):

**Concept**: Region
  **Definition**
    **[EP/NSMC]** A region is a topographical location.
    **[EP/NSIC]** Two regions are the same if and only if they have the same parts.

Example 6:

**Concept**: Respiratory system
  **Definition**
    **[EP/NSMC]** A respiratory system is a physiological system which performs a respiratory function
    **[EP/UC]** The respiratory system is composed of all the organs which contribute to performance of a respiratory function.

## 2.4 Determination of the *structure* of conditions

*2.4.1 Notions*

In order to obtain a model of concepts and relations which is more detailed and thus easier to represent in operational language, OntoSpec recommends explicitating the "structure" of the conditions (for at least that of the membership conditions) in a further step.
  The statements of the following conditions:

  **NMC**: All cars are transportation vehicles.
  **NMC**: All vegetarian eat only fruits and vegetables.
  **NSMC**: All confidential documents are documents whose divulgation outside a defined readership is forbidden.

  effectively refer to properties (underlined) which can be qualified as being either *elementary* (like *being a transportation vehicle*) or *compound* (like *eating only fruits or vegetables* or *being a document whose divulgation outside a defined readership is forbidden*). The terms "elementary" and "compound" are understood here in the same sense as when they are attached to the term "proposition": a property is *elementary* when it is atomic, i.e. not decomposable; a *compound* property is a combination of properties calling on connectors and quantifiers. Considering that a property such as *being a transportation vehicle* is elementary, is equivalent to considering that the syntagm "transportation vehicle" expresses one concept. Conversely, considering that a property such as *only eating fruits and vegetables* is a compound property, is equivalent to considering that the phrase expresses a functional proposition composed (by means of connectors), of the relation named *eat* and of concepts named *Fruits* and *Vegetables*. These are modelling choices. Making these choices leads either to explicit links between concepts and relations already present in the ontology or to the introduction of new concepts and relations into the ontology. In all cases, the result is certainly a finer-grained knowledge model.
  The conceptual entities (concepts and relations) composing an ontology correspond to elementary properties (they are indeed modelled as such). This is typically the case for concepts such as *car*, *transportation vehicle* and *vegetarian*, and for relations such as *eat* and *have for part*. Conversely, as we have just seen, the properties carried by these entities may be arbitrary. OntoSpec recommends specifying the nature of these properties (i.e. their structure), while limiting itself to certain membership conditions. Indeed, on one part, certain conditions – *subsumption links* – play an important role in the structuring of the ontology, whereas, on the other part, some conditions can be represented by means of operational, ontology-dedicated languages.

The simplest NMCs correspond to cases where the carried property is itself elementary, and can thus form part of the ontology. The relation between two properties corresponds to a *subsumption link* (abbreviated as "SL").

Let $\varphi$ and $\psi$ be two distinct properties:

Definition 7: The elementary property $\psi$ *subsumes* the elementary property $\varphi$ if and only if $\varphi$ carries $\psi$ as an NMC.

The equivalent in logic of the statement "$\psi$ subsumes $\varphi$" is the following formula:

$\forall x \; \varphi(x) \rightarrow \psi(x)$; if $\varphi$ and $\psi$ are elementary concepts.
$\forall x_1, x_2 \ldots, x_n \; \varphi(x_1, x_2 \ldots, x_n) \rightarrow \psi(x_1, x_2 \ldots, x_n)$; if $\varphi$ and $\psi$ are n-ary elementary relations ($n \geq 2$).

Comment: One can note that the notion of *subsumption* that we have just defined differs from that in DOLCE, since the latter considers the implication between properties as a *necessarily* true proposition. Again, with the goal of separating the essential from contingent properties, OntoSpec distinguishes between essential and contingent subsumption links.

When the difference between $\varphi$ and $\psi$ is known (we shall refer to it as $\delta$), the existence of such a difference is stated by means of a NSMC. The relation between $\varphi$ and $\psi$ thus becomes a *subsumption link with differentia* (abbreviated as "SLD").

Definition 8: The elementary property $\psi$ *subsumes* the elementary property $\varphi$ *with the differentia* $\delta$ if and only if $\varphi$ carries the property $\psi \wedge \delta$ as an NSMC.

The equivalent in logic of the statement "$\varphi$ subsumes $\psi$ with the differentia $\delta$" is the following formula:

$\forall x \; \varphi(x) \leftrightarrow \psi(x) \wedge \delta(x)$; if $\varphi$ and $\psi$ are elementary concepts and $\delta$ is an arbitrary concept.
$\forall x_1, x_2 \ldots, x_n \; \varphi(x_1, x_2 \ldots, x_n) \leftrightarrow \psi(x_1, x_2 \ldots, x_n) \wedge \delta(x_1, x_2 \ldots, x_n)$; if $\varphi$ and $\psi$ are elementary n-ary relations and $\delta$ is an arbitrary n-ary ($n \geq 2$).

Comment: The fact of considering two properties $\varphi$ and $\psi$ means that they have a semantic content which is similar but not identical, or, in other words, that two different intensions correspond to them. The existence of two types of subsumption links (i.e. with or without differentia), gives the modeller the choice of explicitly stating this difference or not. In the same way, when $\psi$ subsumes two siblings properties $\varphi_1$ and $\varphi_2$, this means that the properties $\varphi_1$ and $\varphi_2$ are similar (since this proximity is shared with $\psi$) but nevertheless different. Here, one encounters the principles of *community* and *difference* with the parent and the siblings stated in the ARCHONTE methodology and which enable one to locally organise the ontology at the single property level [Bachimont, 2004].

The identification of subsumption links has important consequences on the overall structuring of the ontology. Habitually, when the same property is carried by many properties related by means of subsumption links, one expresses this shared property only once, taking into account the "inheritance" of properties. The definitions 9 to 11 below define several inheritance rules, taking into account the various conditions' differing behaviours. The consequence of these definitions is that the shared property becomes tied up to the property which "supplies" this shared property, that is to say the property which carries it for the first time (here we adopt the terminology introduced by [Guarino et Welty, 2000]). In the definitions, the letter O designates the ontology under construction.

<u>Definition 9</u>: A property $\varphi$ of O *supplies* the property $\psi$ of O *as an NMC/NSIC/NIC/SIC/UC* if and only if:
  i) $\varphi$ carries $\psi$ as an NMC/NSIC/NIC/SIC/UC, and
  ii) no property of O subsuming $\varphi$ carries $\psi$ as an NMC/NSIC/NIC/SIC/UC.

<u>Comment</u>: In the case of the identity and unity conditions, the properties $\varphi$ and $\psi$ are only concepts.

<u>Definition 10</u>: A property $\varphi$ of O *supplies* the property $\psi$ of O *as an SMC* if and only if:
  i) $\varphi$ carries $\psi$ as an SMC, and
  ii) no property of O subsumed by $\varphi$ carries $\psi$ as an SMC.

<u>Definition 11</u>: A property $\varphi$ of O *supplies* the property $\psi \wedge \delta$ *as an NSMC* ($\psi$ being a property of O) if and only if:
  i) $\varphi$ supplies $\psi \wedge \delta$ as an SMC, and
  ii) $\varphi$ carries $\delta$ as an NMC, and
  ii) $\psi$ does not carry $\delta$ as an NMC.

<u>Comment</u>: The conjunction of conditions ii) and iii) does not (logically) imply that $\varphi$ carries $\delta$ as an NMC. Indeed, in the event of multiple inheritance, $\varphi$ can be subsumed by another concept (or relation) not subsuming $\psi$ and which carries $\delta$ as an NMC.

In addition to subsumption links, OntoSpec recommends identifying other kinds of NMCs: the very NMCs that are taken into account by web ontology languages in general and the OWL language in particular [Antoniou and van Harmelen, 2004][13].

For concepts, OntoSpec notably recommends identifying Existential Restrictions (ER), Value Restrictions (VR)[14] and Incompatibility Links (ICL) as NMCs. Below, we give the logical equivalents of these NMCs.

Let $\varphi$ and $\varphi'$ be two elementary concepts and $\psi$ an elementary relation.

Existential Restriction: $\forall x\ \varphi(x) \rightarrow \exists y(\varphi'(y) \wedge \psi(x,y))$
Value Restriction: $\forall x\ \varphi(x) \rightarrow \forall y(\psi(x,y) \rightarrow \varphi'(y))$

---

[13] The OWL-DL dialect integrates Description Logics constructors such as "existential restrictions" and "value restrictions". In OntoSpec, these correspond to NMC categories bearing the same name, so as to facilitate subsequent translation of OntoSpec-specified ontologies into OWL.
[14] The value restrictions are also called "universal restrictions" (represented in OWL by the constructor *allValuesFrom*), as opposed to the existential restrictions (represented in OWL by the constructor *someValuesFrom*).

Incompatibility Link: $\forall x\ \varphi(x) \rightarrow \neg\varphi'(x)$

For relations, OntoSpec notably recommends identifying Domain Restrictions (DR), Range Restrictions (RR) and Inverse Links (IVL) as NMCs.

Let $\varphi$ and $\varphi'$ be two elementary binary relations and $\psi$ an elementary concept.

Domain Restriction: $\forall x,y\ \varphi(x,y) \rightarrow \psi(x)$
Range Restriction: $\forall x,y\ \varphi(x,y) \rightarrow \psi(y)$
Inverse Link: $\forall x,y\ \varphi(x,y) \leftrightarrow \varphi'(y,x)$

In Appendix 1 of this report, we give a complete list of conditions that OntoSpec recommends identifying, along with their respective acronym.

Comments:

- The identification and the distinction of these categories of NMC necessitates good command of the semantics of the equivalent logical formulae in general and the meaning of the connectors $\wedge$ and $\rightarrow$ in particular. However, analysis of current practice in construction of formal ontologies in OWL[15] shows that most modellers encounter difficulties in this respect [Rector *et al.*, 2004]. This observation has prompted us to anticipate the provision of assistance to the modeller so that he/she can complete this step successfully. At present, OntoSpec lacks this type of help.
- Furthermore, OntoSpec does not currently request specification of the structure of the subsumption links with differentia: for example, the difference $\delta$ may take the form of an existential or value restriction. However, we do plan to provide these extensions.

*2.4.2 Examples*

Examples 7 and 8 (below) show that the acronyms associated with the more specific NMCs substitute for the acronym NMC in the property labels. It should be noted that the indication of the type of NMC enables the expression of a property not be a simple natural language paraphrase of the logical content: thus, for the third property of the concept *all-terrain car* it is not stated that "if an all-terrain car possesses wheels, then these wheels are necessarily drive wheels". One should also note that the second property of the relation *eat* is the conjunction of a domain restriction with a range restriction.

Again, we use these examples to specify a typographic rule used in OntoSpec: when a word or syntagm corresponds to the label of a concept or a relation, it is respectively noted in capital letters (e.g. CAR, RAISED CHASSIS) or in italics (e.g. *possesses*, *eats*) in order to refer to this conceptual entity within the ontology. This typographic rule - adopted and adapted from the expression mode in Enterprise Ontology [Uschold *et al.*, 1998] – allows one to indicate that the word or syntagm is used in a precise sense. Otherwise, the word or syntagm (e.g. drives, terrain) is to be understood in an everyday sense – that given in a dictionary, for example.

---

[15] Of course, this analysis goes beyond the case of OWL alone and concerns all formal languages with logical structures.

Example 7:

**Concept**: All-terrain car
  **Definition**
   **[EP/SLD]** An ALL-TERRAIN CAR is a CAR which allows one to drive over all types of terrain.
   **[EP/ER]** All ALL-TERRAIN CARS *possess* a RAISED CHASSIS.
   **[EP/VR]** All ALL-TERRAIN CARS *possess* only DRIVE WHEELS.

Example 8:

**Relation**: eat
  **Definition**
   **[EP/SLD]** *to eat* is *to feed oneself*.
   **[EP/DR & RR]** A LIVING BEING eats SOLID FOODS.

## 2.5 Attribution of *meta-properties*

### 2.5.1 Notions

OntoSpec enables one to complete the definition of conceptual entities by attributing them with meta-properties. The prefix "meta" means that these properties bear on the conceptual entities themselves, and not on the instances of these conceptual entities. The domain of discourse is thus different.

Common examples of meta-properties that apply to binary relations are the meta-properties of "reflexivity", "symmetry" and 'transitivity". Stating that a relation *rel* is symmetric, for example, can be done in two ways: either by attributing a property to the couples in the relation ($\forall x,y\ rel(x,y) \rightarrow rel(y,x)$), or by declaring that "the relation *rel* is symmetric". Since the notion of "symmetric" corresponds to the previous schema of axioms (which can be instantiated by any relation *rel*), one must consider that the two statements are equivalent. Attributing a meta-property is useful because it allows more synthetic statements. In the present case, one can assimilate this possibility to a syntactical "sweetener".

Examples of meta-properties, applying now to concepts, are the meta-properties "defined" and "primitive", which amount to noting the presence or absence of NSMCs in the definition of the concept.

Definition 12: A concept is *defined* (respectively *primitive*) if and only if it carries (respectively does not carry) an NSMC.

This categorisation turns out to be important at the time of formalisation because defined and primitive formal concepts have different behaviours with regard to the inferential services associated with operational languages: for example, only defined concepts are taken into account by the concept classifiers associated with Descriptions Logics. As in the case of the symmetry for a relation, these meta-properties can be considered as syntactical sweeteners. One should note however that meta-properties offer additional convenience: it is possible to attribute them to a concept without having to display an NSMC.

In addition to providing the ability (in general) to express meta-properties, OntoSpec incites the designer to attribute certain meta-properties in particular, i.e., those on which the OntoClean methodology is based [Guarino and Welty, 2002, 2004]: In fact, OntoSpec incorporates OntoClean, which contributes to ontology construction in the following manner:

on the basis of the attribution of certain meta-properties to concepts, the designer possesses rules for evaluating the logical consistence of the subsumption links he/she has modelled.

As in the case of "defined" and "primitive" meta-properties, a first set of meta-properties accounts for the existence of a given type of condition carried or supplied by the concept. These are summarized below, together with a reminder of the abbreviated notation proposed by OntoClean (e.g. +**I**, -**R**) and that we have reused in OntoSpec:

- The concept carries (respectively does not carry) a common IC. The latter can be necessary and sufficient or simply necessary or sufficient. Notations: +**I** (respectively -**I**).
- The concept supplies a common IC. Notation: +**O**.
- The concept carries (respectively does not carry) a common UC: +**U** (respectively -**U**).
- The concept does not carry a common UC for any of its instances: ~**U**.

A second set of meta-properties accounts for the modal behaviour of concepts. In the modal logic formulae used in definitions 13 to 15, we use "nec" to refer to the necessity operator[16].

<u>Definition 13</u>: A concept $\phi$ is *rigid* if and only if it is essential for all its instances, i.e. it is such that: $\forall x\ \phi(x) \rightarrow nec\ \phi(x)$. Notation: +**R**.

<u>Definition 14</u>: A concept $\phi$ is *non-rigid* if and only if it is not essential for some of its instances, i.e. it is such that: $\exists x\ \phi(x) \land \neg\ nec\ \phi(x)$. Notation: -**R**.

<u>Definition 15</u>: A concept $\phi$ is *anti-rigid* if and only if it is not essential for all of its instances, that is if it is such that: $\forall x\ \phi(x) \rightarrow \neg\ nec\ \phi(x)$. Notation: ~**R**.

<u>Comments</u>:
- The notion of *rigidity* can be likened to that of a property being *essential* with regard to another property (cf. definition 1). In a way, a rigid concept can be classified as an "intrinsically" essential concept.
- According to these definitions, an *anti-rigid* concept is *non-rigid*.

Finally, one other meta-property, of "dependence", allows one to account for the fact that the existence of an entity necessarily implies the existence of another entity. Several types of dependence can be defined, and this is effectively the case in the DOLCE ontology. The notion of dependence chosen for OntoClean is that of a generic dependence on an external entity. In the definition below, the predicates *P* and *K* represent the atemporal relations *all-parts* and *constitutio*n respectively.

<u>Definition 16</u>: A concept $\phi$ is *externally dependent on* a concept $\psi$, or *carries* the concept $\psi$ *as an External Dependency Condition* (EDC) if and only if for each instance x of $\phi$ there is necessarily an instance y of $\psi$ which is neither a part nor a constituent of x, that is to say: $\forall x\ nec(\phi(x) \rightarrow \exists y\ \psi(y) \land \neg P(y,x) \land \neg K(y,x))$. Notations: a concept is dependent (+**D**) or not (-**D**).

---

[16] A more complex formulation of these definitions (taking time into account) has been used on various occasions in papers by Guarino and Welty. More over, improvements in these definitions have recently been proposed by other authors [Andersen and Menzel, 2004][Carrara *et al.*, 2004]. This situation shows that these notions (and therefore the OntoClean methodology) are still evolving. In this report, we have chosen to use the simplest form. Indeed, our intention is only to specify at which step of the OntoSpec methodology these notions must be taken into account and what the notions' roles are.

Taking into account the inheritance of properties, via subsumption links, one can use the previous definitions to deduce a certain number of rules concerning the presence within the definitions of supplied (or merely carried) properties. We call these rules *Subsumption Constraints* (SC), as in the OntoClean methodology, meaning that they must be satisfied each time the builder introduces a new subsumption link into the ontology.

Let φ and φ' be two concepts in the ontology and let ψ be an arbitrary property.

<u>SC 1</u>: **If** φ supplies (or simply carries) ψ as an NMC/IC/UC/EDC **and** φ subsumes φ', **then** φ' carries ψ as an NMC/IC/UC/EDC.

<u>SC 2</u>: **If** φ supplies (or simply carries) ψ as an SMC **and** φ' subsumes φ, **then** φ' carries ψ as an SMC.

<u>SC 3</u>: **If** φ does not carry a common UC for any of its instances (anti-unity) **and** φ' subsumes φ, **then** φ' does not carry a common UC for any of its instances.

<u>SC 4</u>: **If** φ is anti-rigid **and** φ' subsumes φ, **then** φ' is anti-rigid.

*2.5.2 Examples*

Example 9 specifies the syntax used to state meta-properties. The key-words *Meta-properties* and *Properties* substitute for the key-word *Definition*. The same typographic conventions apply to expression of meta-properties, in particular for what concerns the labels of meta-concepts (noted in capital letters). When these meta-properties correspond meta-properties in OntoClean, the abbreviations recommended by the latter are used.

<u>Example 9</u> (continued from Example 7):

**Concept**: All-terrain car
  **Meta-properties**
    ALL-TERRAIN CAR is DEFINED. ALL-TERRAIN CAR is RIGID (+**R**). ALL-TERRAIN CAR CARRIES AN IDENTITY CONDITION (+**I**). ALL-TERRAIN CAR CARRIES A UNITY CONDITION (+**U**). ALL-TERRAIN CAR is NON-DEPENDENT (-**D**).
  **Properties**
    **[EP/SLD]** An ALL-TERRAIN CAR is a CAR which allows one to drive over all type of terrain.
    **[EP/ER]** All ALL-TERRAIN CAR *possess* a RAISED CHASSIS.
    **[EP/VR]** All ALL-TERRAIN CAR *possess* only DRIVE WHEELS.

Example 10 illustrates the contribution of the notions of *rigidity* and *dependence* to the characterisation of concepts. The anti-rigidity attributed to the concept *Student* means that no student is supposed to remain a student along all his/her life (this property is contingent for all instances of the concept). The dependence signifies that the status of being a student is conferred by a higher education institution, and that the former would not exist without the latter.

Example 10:

**Concept**: Student
  **Meta-properties**
   STUDENT is DEFINED. STUDENT is ANTI-RIGID (~**R**). STUDENT CARRIES AN IDENTITY CONDITION (+**I**). STUDENT CARRIES A UNITY CONDITION (+**U**). STUDENT is DEPENDENT (+**D**).
  **Properties**
   [EP/SLD] A STUDENT is a PERSON who studies in higher education.
   [EP/ER] All STUDENTS *are registered in* a HIGHER EDUCATION INSTITUTION.

## 2.6 Structuring of comments

As the structuring of the definition of a conceptual entity proceeds, comments are assembled in a separate text. In order to facilitate reading of this text, OntoSpec identifies certain recurrent types of comments, corresponding (for example) to the addition of examples or counter-examples, specification of a community of meaning (semantic axis) within the subsumed entities or even provision of complementary information by means of citations.

In this section, we shall merely provide a few examples illustrating the use of these different comments. The *Person* concept illustrates the "semantic axis" [**SA**] comment and the *Endurant* concept illustrates the "examples" [**EX**], "counter-examples" [**CEX**] and "citation" [**CIT**] comments. Then *Watercourse* concept illustrates the "other" comment, noted here as "diverse" [**DIV**].

Example 11:

**Concept**: Person
  **Meta-Properties**
   PERSON is PRIMITIVE. PERSON is RIGID (+**R**). PERSON SUPPLIES AN IDENTITY CONDITION (+**O**). PERSON CARRIES A UNITY CONDITION (+**U**). PERSON is NON-DEPENDENT (-**D**).
  **Properties**
   [**EP/SL**] A PERSON is an ANIMATE.
   [**EP/ER**] Every PERSON *is constituted by* a BIOLOGICAL BODY.
   [**CP/ER**] Every PERSON *possesses* a SECURITY SOCIAL NUMBER.
  **Comment**
   [**SA**] The concept PERSON is refined into MAN and WOMAN according to the relation *is of sex*.

Example 12:

**Concept**: Endurant
  **Meta-Properties**
   ENDURANT is PRIMITIVE. ENDURANT is RIGID (+**R**). ENDURANT does not CARRY AN IDENTITY CONDITION (-**I**). ENDURANT does not CARRY A UNITY CONDITION (-**U**). ENDURANT is NON-DEPENDENT (-**D**).
  **Properties**
   [**EP/SL**] An ENDURANT is a PARTICULAR.
  **Comment**

[**SA**] The ENDURANTS are divided into PHYSICAL ENDURANTS and NON-PHYSICAL ENDURANTS according to whether or not they have direct spatial qualities.
[**EX**] An apple, a person or an idea are examples of ENDURANTS.
[**CEX**] The fall of an apple, the birth of a person or the genesis of an idea are counter-examples of ENDURANTS, being considered as PERDURANTS generated by ENDURANTS.
[**CIT**] [D18, p. 15] "Endurants are *wholly* present (i.e. all their proper parts are present) at any time they are present."
[**CIT**] [D18, p. 16] "Endurants can "genuinely" change in time, in the sense that the very same endurant as a whole can have incompatible properties at different times."

Example 13:

**Concept**: Watercourse
  **Meta-Properties**
  WATERCOURSE is DEFINED. WATERCOURSE is RIGID (+**R**). WATERCOURSE SUPPLIES AN IDENTITY CONDITION (+**0**). WATERCOURSE CARRIES A UNITY CONDITION (+**U**). WATERCOURSE is NON-DEPENDENT (-**D**).
  **Properties**
  [**EP/SLD**] A WATERCOURSE is a CHANNEL which is followed by a SURFACE WATER FLOW.
  **Commentaires**
  [**DIV**] The SURFACE WATER FLOW originates notably from run-off, springs, melting snow, a pond or water-logged areas.

## 3 Re-ingineering of DOLCE in OntoSpec

The re-engineering process has primarily consisted in translating the logical theory representing DOLCE (which is completely presented in D18, section 4) into the semi-informal language of OntoSpec. We describe this process below by following the different steps recommended by the OntoSpec methodology. We refer to the resulting ontology as "DOLCE-OS". The latter is fully presented in Appendix 2.

**3.1 Structuring of the ontology**

The first step in OntoSpec consists in identifying the conceptual entities (concepts and relations) and, for each of them, in structuring the definition by distinguishing a statement of the properties satisfied by the instances (the "definition" *per se*) on one hand, and a comment on the other.
  This step is simplified in the present case, given that the initial data already corresponds to an ontology. In particular, the list of conceptual entities is set. DOLCE presents itself as an ontology of "particulars" (the name given to the ontology's quantification domain) and is constituted by a certain number of concepts and relations which are instantiated by particulars:
- 37 rigid concepts (constituting the set named $\Pi_X$) and 1 non-rigid concept, the concept named *Atomic*.
- 29 binary relations and 21 ternary relations.

One should note that DOLCE also defines concepts and relations which deal with properties rather than particulars. These meta-properties include (for example) the meta-concept *Rigid* and the meta-relation *subsumes*, as well as meta-relations modelling different kinds of dependence between concepts. Most of these meta-properties are re-used in DOLCE-OS, and are specified in a semi-informal way (so as to be homogeneous with the other definitions): the latter allow one to attribute meta-properties to the concepts within the ontology of particulars.

As for the distinction between definition and comment, here again things are relatively simple. Since DOLCE is an axiomatic theory, translation of the axioms constitutes the definition of the conceptual entities. According to the form taken by the axioms, the latter are distributed across the various definitions of concepts or relations. Otherwise, the numerous informal comments (corresponding to clarifications, examples and references to other works) present in section 3 of D18 (introducing DOLCE's content) are reused, so as to constitute the comments for the definitions in DOLCE-OS.

### 3.2 Distinction between essential and contingent properties

Since all of DOLCE's axioms and theorems are *necessarily* true, all of DOLCE-OS's properties are essential.

### 3.3 Explicitation of the categories of conditions

At a first level for structuring the definitions, OntoSpec requires identification of membership, identity and unity conditions, for concepts, but solely membership conditions for relations. On the whole, membership conditions are acquired from axiomatic theory, subject to reformulation of the formulae. Identity and unity conditions are obtained from informal comments accompanying the axioms.

Examination of DOLCE's formulae (definitions, axioms, theorems), such as:
- (Dd14) $PP(x,y) =_{def} P(x,y) \land \neg P(y,x)$
- (Ad2) $P(x,y) \rightarrow (PD(x) \leftrightarrow PD(y))$

prompts us to ask the following question: which conceptual entities do the formulae help define? In other words: in which definitions should they appear?

In the case of (D14), the answer is straightforward: this formula expresses a necessary and sufficient membership condition for the relation PP (Proper Part). In the case of (Ad2), the implication may suggest a necessary membership condition for the relation P (Part). However the paraphrase in English: "a perdurant is a part of perdurants only, and has for parts only perdurants" rather suggests a constraint bearing on the notion of perdurant. Indeed, the axiom (Ad2) is logically equivalent to the conjunction of the axioms (Ad2a) and (Ad2b) below, which can be considered as necessary membership conditions for the concept PD (Perdurant):
- (Ad2a) $PD(x) \rightarrow (P(y,x) \rightarrow PD(y))$
- (Ad2b) $PD(x) \rightarrow (P(x,y) \rightarrow PD(y))$

Below, we give other representative examples of the formula transformations that we have performed. The general idea is to eliminate the conjunctions in order to achieve simpler propositions. The link to a conceptual entity then becomes natural. We notably find (see below) that (Ad5a)(Ad5b)(Ad8a) and (Ad8b) express necessary membership conditions for the concepts AB (Abstract) and PD (Perdurant). In the case of (TD1), we have taken into account the signature of the relation K (constitution) – where the argument x is necessarily an ED (Endurant) or a PD (Perdurant) – to make the membership conditions appear clearly. With the addition of an apostrophe, the notation of the theorems (Td1a') and (TD1b') indicates that the theorems have been obtained from (Td1) – to ensure traceability – but that they are not logically equivalent.
- (Ad5) $(AB(x) \lor PD(x)) \to P(x,x)$
    - (Ad5a) $AB(x) \to P(x,x)$
    - (Ad5b) $PD(x) \to P(x,x)$
- (Ad8) $((AB(x) \lor PD(x)) \land \neg P(x,y)) \to \exists z\, (P(z,x) \land \neg O(z,y))$
    - (Ad8a) $AB(x) \to (\neg P(x,y) \to \exists z\, (P(z,x) \land \neg O(z,y)))$
    - (Ad8b) $PD(x) \to (\neg P(x,y) \to \exists z\, (P(z,x) \land \neg O(z,y)))$
- (Td1) $\neg K(x,x,t)$
    - (Td1a') $ED(x) \to \neg K(x,x,t)$
    - (Td1b') $PD(x) \to \neg K(x,x,t)$

Once these transformations have being carried out by paraphrasing the formulae in natural language, we obtain definitions such as that in Example 14. One can note that in DOLCE-OS we maintain a link with the translated formula by indicating its name in square brackets. Again with respect to Example 14, it is also noteworthy that in the paraphrase of the axiom (Ad8) (corresponding to the last property) we have specified that the variables x, y and z are necessarily perdurants (they are named *perdurant1*, *perdurant2* and *perdurant3*) in order to make the statement easier to read. Taking this addition into account, the reference axiom is denoted by an apostrophe at the end (Ad8').

Example 14:

**Concept**: Perdurant
  **Properties**
    [EP/NMC] A perdurant is a particular.
    [**Ad2a**; EP/NMC] Each perdurant has for parts only perdurants.
    [**Ad2b**; EP/NMC] Each perdurant is part of perdurants only.
    [**Ad5b**; EP/NMC] Each perdurant is part of itself.
    [**Ad8b'**; EP/NMC] Every perdurant1 which is not part of a perdurant2 is such that at least one perdurant3 exists which is part of perdurant1 and which does not overlap with perdurant2.

**3.4 Specifying the structure of the conditions**

Here, it is a matter of describing the structure of the membership conditions (while displaying links with other conceptual entities) and identifying certain categories of conditions (subsumption links, existential restrictions, etc.). Syntactically, reference to other conceptual entities is performed by noting their name in capital letters (for the concepts) and in italics (for relations).

In Example 15 (which adopts its definition of the concept of endurant from Example 14), we show the result of this step. One subsumption link (SL) and two value restrictions (VR) have been identified. One can note that the axiom (Ad2) has been translated into the axiom (Ad2'). The reason is that in order to facilitate the expression and therefore comprehension of the property, one must refer to the relation *has for part*, the inverse of the relation *is part of* (this aspect already appears in Example 14). In order to obtain such expressions, some inverse relations have been added in DOLCE-OS.

Example 15 (continued from Example 14):

**Concept**: Perdurant
  **Properties**
   [EP/SL] A PERDURANT is a PARTICULAR.
   [**Ad2a'**; EP/VR] A PERDURANT *has for parts* only PERDURANTS.
   [**Ad2b**; EP/VR] A PERDURANT *is part of* PERDURANTS only.
   [**Ad5b**; EP/NMC] Each PERDURANT *is part of* itself.
   [**Ad8b'**; EP/NMC] Each PERDURANT1 which *is not part of* a PERDURANT2 is such that at least one PERDURANT3 exists which *is part of* PERDURANT1 and which does not *overlap with* PERDURANT2.

**3.5 Attribution of meta-properties**

DOLCE-OS's concepts are characterized by two categories of meta-properties: DOLCE-specific meta-properties (defined in D18 by means of logical meta-theory), on one hand, and meta-properties corresponding to application of the OntoClean methodology, on the other.

The first category notably includes the property of being *non-empty* (NEP), that is to necessarily admitting instances, but also the property whereby a given set of concepts constitutes a *partition* of a sub-domain of instances (PT) and, equally, various dependence relations: in particular, *qualities* depend (mutually) on the entities in which they are inherent, *perdurants* depend on *endurants* which generate them and *endurants* depend on their *constituents* (in the sense of the K *constitution* relation), the latter ultimately being a *quantity of matter* (M). Finally, all concepts (except for *Atomic*) are considered as *rigids*, this last meta-property being shared by the OntoClean methodology.

OntoClean-specific meta-properties also include an overall summary of the existence of identity and unity conditions. One should note that unity conditions do not concern *qualities* (Q) because no parthood relation is defined for these entities. Finally, external dependence links are stated - for example, the fact that a *perdurant* depends externally on an *endurant*. Since this relation is more constrained than the dependence relations defined in DOLCE, one should not be surprised to see that some concepts are defined as being non-dependent in OntoClean's sense (notation: -D), whereas dependence links with external entities are indeed stated (in particular, as we already have seen, every endurant or perdurant in DOLCE depends on its qualities).

Example 16 (continued from Example 15):

**Concept** Perdurant
**Meta-properties**
  PERDURANT is RIGID (+**R**). PERDURANT is NOT CARRYING AN IDENTITY CRITERION (-**I**). PERDURANT is NOT CARRYING A COMMON UNITY CRITERION (-**U**). PERDURANT is EXTERNALLY-DEPENDENT (+**D**). PERDURANT *mutually specifically constantly depends on* TEMPORAL QUALITY. PERDURANT *inversely partially generically spatially depends on* ENDURANT. PERDURANT is NON-EMPTY. EVENT and STATIVE *is a non-trivial Partition of* PERDURANT.
**Properties**
  [EP/SL] A PERDURANT is a PARTICULAR.
  [**Ad2a'**; EP/VR] A PERDURANT *has for parts* only PERDURANTS.
  [**Ad2b**; EP/VR] A PERDURANT *is a part of* PERDURANTS only.
  [**Ad5b**; EP/NMC] Every PERDURANT *is part of* itself.

[**Ad8b'**; EP/NMC] Every PERDURANT1 which *is* not *a part of* a PERDURANT2 is such that at least one PERDURANT3 exists which *is a part of* PERDURANT1 and which does not *overlap with* PERDURANT2.

## 4 Conclusion

As a result of our work, DOLCE-OS constitutes a resource for the OntoSpec methodology. Its role is to help builders construct application ontologies at the "knowledge level". Firstly DOLCE-OS delivers generic modelling principles aiming at structuring the conceptualisation of any application domain. These principles include (for example) the distinction between *endurants* and *perdurants* and the fact of considering *qualities* as individuals which are *inherent* to the entities they qualify, i.e. both being specific for these entities (endurants and perdurants) and having the same temporal extension [Masolo and Borgo, 2005]. Secondly, this resource implements the combined definition principles of OntoSpec and OntoClean. In this respect, one can note that the implementation of these latter principles does not require the use of a formal knowledge representation language (as was already exemplified by the integration of OntoClean in Methontology [Fernández-López and Gómez-Pérez, 2002])

One outcome of this work is that DOLCE-OS equally constitutes a "showcase" for the OntoSpec methodology, because the former completely illustrates the latter's semi-informal specification mode. Hence, from now on, builders can advantageously employ DOLCE-OS to help master the OntoSpec language.

We must now evaluate in practice DOLCE-OS's contribution to ontology construction projects. In this respect, we wish to emphasize that the OntoSpec methodology's recent integration into the TERMINAE plat-form, in the form of a "modelling files editor" [Bruaux, 2005] should provide us with room for experiment.

Returning to DOLCE, we can also mention some of our current development perspectives. Our objective is thus to integrate extensions of the original DOLCE kernel into DOLCE-OS. For example, DOLCE+ (D18, chapter 15) extends DOLCE in several directions by integrating subontologies currently being studied at the LOA - in particular an ontology of descriptions and situations called D&S [Gangemi and Mika, 2003] and a computational ontology of mind [Ferrario and Oltramari, 2004]). We at the LaRIA have elaborated (again as extensions of DOLCE) an ontology of organisations and an ontology dedicated to document content analysis [Fortier, 2005]. Moreover, we are working on the design of a problem-solving core ontology [Bruaux *et al.*, 2005]. Our mid – and long term ambition is to radically change the way in which application ontologies are built by ensuring that this task mainly involves the reuse and adaptation of existing core and application ontologies.

## Thanks

We thank Sylvie Després and Sylvie Szulman whose remarks on a first draft of the text enabled us to make substantial improvements.

## References

W. Andersen and C. Menzel (2004). Modal Rigidity in the OntoClean Methodology. In A.C. Varzi and L. Vieu (eds.), *Formal Ontology in Information Systems*, Proceedings of the *International Conference FOIS 2004*, IOS Press, p. 119-127, November 2004.

G. Antoniou and F. van Harmelen (2004). Web Ontology Language: OWL. In S. Staab and R. Studer (eds.), *Handbook on Ontologies*, Springer Verlag, p. 67-92, 2004.

N. Aussenac-Gilles, B. Biébow and S. Szulman (2000). Revisiting ontology design: a methodology based on corpus analysis. In R. Dieng and O. Corby (eds.), *Knowledge Engineering and Knowledge Management: Methods, Models and Tools*, Proceedings of the *12th International Conference EKAW'2000*, Springer-Verlag, LNAI 1937, p. 172-188, 2000.

B. Bachimont (2004). Art et sciences du numérique : ingénierie des connaissances et critique de la raison computationnelle. HDR (Habilitation à Diriger des Recherches) report, University of Technology of Compiègne, January 2004.

S. Bechhofer, I. Horrocks, C. Goble and R. Stevens (2001). OilEd: a reason-able ontology editor for the Semantic Web. In Proceedings of the *Joint German/Austrian Conference on Artificial Intelligence: KI'01*, LNAI 2174, Springer Verlag, p. 396-408, 2001.

S. Ben Khédija (2004). Intégration de la méthode OntoSpec dans TERMINAE. DEA (Diplôme d'Études Approfondies) report, University of Picardie Jules Verne, Amiens, July 2004. *Also published as a technical report from the ATONANT project*.

T. Bittner, M. Donnelly and B. Smith (2004). Individuals, Universals, Collections: On the Foundational Relations of Ontology. In A.C. Varzi and L. Vieu (eds.), *Formal Ontology in Information Systems*, Proceedings of the *Third International Conference: FOIS-2004*, IOS Press, p. 37-48, November 2004.

S. Bruaux. (2005). Intégration d'un éditeur de fiches de modélisation dans la plate-forme TERMINAE. Technical report from the ATONANT project, September 2005.

S. Bruaux, G. Kassel and Gilles Morel (2005). An ontological approach to the construction of problem-solving models. LaRIA Research Report 2005-03, University of Picardie Jules Verne. Available at <http://hal.ccsd.cnrs.fr/ccsd-00005019>.

M. Carrara, P. Giaretta, V. Morato, M. Soavi and G. Spolaore (2004). Identity and Modality in OntoClan. In A.C. Varzi and L. Vieu (eds.), *Formal Ontology in Information Systems*, Proceedings of the *International Conference FOIS 2004*, IOS Press, p. 128-139, November 2004.

M. Fernández-López, A. Gómez-Pérez, J. Pazos-Sierra and A. Pazos-Sierra (1999). Building a Chemical Ontology Using Methontology and the Ontology Design Environment. *IEEE Intelligent Systems*, p. 37-46, January/February 1999.

M. Fernández-López and A. Gómez-Pérez (2002). The Integration of OntoClean in WebODE. In Proceedings of the *EKAW 2002 Workshop on Evaluation of Ontology-based Tools (EON2002)*, Siguenza (Spain), 2002.

R. Ferrario and A. Oltramari (2004). Towards a Computational Ontology of Mind. In A.C. Varzi and L. Vieu (eds.), *Formal Ontology in Information Systems*, Proceedings of the *International Conference FOIS 2004*, IOS Press, p. 287-297, November 2004.

J.-Y. Fortier (2005). Vers une gestion des connaissances au niveau des informations. PhD thesis, University of Picardie Jules Verne, October 2005.

A. Gangemi and P. Mika. Understanding the Semantic Web through Descriptions and Situations. In R. Meersman *et al.* (eds), Proceedings of the *International Conference on Ontologies, Databases and Applications of Semantics* (*ODBASE 2003*), Catania (Italy), November 2003.

N. Guarino and P. Giaretta (1995). Ontologies and Knowledge Bases; Towards a terminological Clarification. In N. Mars (ed.), *Towards Very Large Knowledge Bases: Knowledge Building and Knowledge Sharing*, Amsterdam, IOS Press, p. 25-32, 1995.

N. Guarino and C. Welty (2000a). Identity, Unity, and Individuality: Towards a formal toolkit for ontological analysis. In W. Horn (ed.), Proceedings of the *14th European Conference on artificial Intelligence: ECAI-2000*, Berlin: IOS Press, p. 219-223, August 2000.

N. Guarino and C. Welty (2000b). A Formal Ontology of Properties. In R. Dieng and O. Corby (eds.), Proceedings of the *12th International Conference on Knowledge Engineering and Knowledge Management: EKAW-2000*, Lecture Notes on Computer Science, Springer Verlag, p. 97-112, October 2000.

N. Guarino and C. Welty (2002). Evaluating Ontological Decisions with OntoClean. *Communication of the ACM*, 45(2):61-65, February 2002.

N. Guarino and C. Welty (2004). An Overview of OntoClean. In S. Staab and R. Studer (eds.), *Handbook on Ontologies*, Springer Verlag, p. 151-171, 2004.

G. Kassel (2002). OntoSpec : une méthode de spécification semi-informelle d'ontologies. In Proceedings of the *13th French-speaking Conference on Knowledge Engineering: IC'2002*, Rouen (France), p. 75-87, 2002.

C. Masolo, S. Borgo, A. Gangemi, N. Guarino, A. Oltramari and L. Schneider (2003). The WonderWeb Library of Foundational Ontologies and the DOLCE ontology. WonderWeb Deliverable D18, Final Report (vr. 1.0, 31-12-2003).

C. Masolo and S. Borgo (2005). Qualities in Formal Ontology. In proceedings of the *Workshop on Foundational Aspects of Ontologies: FOnt 2005*, Koblenz (Germany), September 2005.

I. Niles and A. Pease (2001). Towards a standard upper ontology. In Proceedings of the *2nd International Conference on Formal Ontology in Information Systems: FOIS-2001*, Ogunquit, Main, October 2001.

N.F. Noy and D.L. McGuinness (2001). Ontology Development 101: A Guide to Creating Your First Ontology. Stanford Knowledge Systems Laboratory, Technical Report KSL-01-05, and Stanford Medical Informatics Technical Report SMI-2001-0880, March 2001.

A. Rector, N. Drummond, M. Horridge, J. Rogers, H. Knublauch, R. Stevens, H. Wang and C. Wroe (2004). OWL Pizzas: Practical Experiences of Teaching OWL-DL: Common Errors and common Patterns. In E. Motta, N. Shadbolt, A. Stutt and N. Gibbins (eds.), *Engineering Knowledge in the Age of the Semantic Web*, Proceedings of the *14th International Conference EKAW'2004*, Whittlebury Hall (UK), p. 63-81, 2004.

S. Szulman and B. Biébow (2004). OWL et Terminae. In Proceedings of the *15th French-speaking Conference on Knowledge Engineering: IC'2004*, Lyon (France), p. 41-52, 2004.

M. Uschold and M. Grüninger (1996). Ontologies: Principles, methods and applications. *Knowledge Engineering Review*, 11(2):93-136, 1996.

M. Uschold, M. King, S. Moralee and Y. Zorgios (1998). The Enterprise Ontology. *Knowledge Engineering Review*, Special Issue on Putting Ontologies to Use, M. Uschold and A. Tate (eds.), vol 13(1):31-89, 1998.

C. Welty and N. Guarino (2001). Supporting Ontological Analysis of Taxonomic Relationships. *Data and Knowledge Engineering*, 39(1):51-74, 2001.

# Appendix 1: List of conditions and their acronym

## A1.1 Conditions characterizing concepts

cpt: concept of the ontology

- Necessary Membership Condition (**NMC**):
  $\forall x\ cpt(x) \rightarrow \psi(x)$; $\psi$: arbitrary concept

  o Subsumption Link (**SL**):
     $\forall x\ cpt(x) \rightarrow cpt'(x)$; cpt': concept of the ontology

  o Existential Restriction (**ER**):
     $\forall x\ cpt(x) \rightarrow \exists y(cpt'(y) \wedge rel(x,y))$; cpt': concept of the ontology; rel: binary relation of the ontology
     $\forall x\ cpt(x) \rightarrow \exists y,z(cpt'(y) \wedge cpt''(z) \wedge rel(x,y,z))$; cpt', cpt'': concepts of the ontology; rel: ternary relation of the ontology

  o Value Restriction (**VR**):
     $\forall x\ cpt(x) \rightarrow \forall y(rel(x,y) \rightarrow cpt'(y))$; cpt': concept of the ontology; rel: binary relation of the ontology
     $\forall x,z\ cpt(x) \rightarrow \forall y(rel(x,y,z) \rightarrow cpt'(y))$; cpt': concept of the ontology; rel: n-ary relation ($n \geq 3$) of the ontology; z: vector of variables

  o Extended Value Restriction (**EVR**):
     $\forall x\ cpt(x) \rightarrow \forall y(rel(x,y) \rightarrow \psi(y))$; $\psi$: arbitrary concept; rel: binary relation of the ontology
     $\forall x,z\ cpt(x) \rightarrow \forall y(rel(x,y,z) \rightarrow \psi(y))$; $\psi$: arbitrary concept; rel: n-ary relation ($n \geq 3$) of the ontology; z: vector of variables

  o Constant Restriction (**CR**):
     $\forall x\ cpt(x) \rightarrow rel(x,a)$; rel: binary relation of the ontology

  o Incompatibility Link (**ICL**):
     $\forall x\ cpt(x) \rightarrow \neg cpt'(x)$; cpt': concept of the ontology

- Sufficient Membership Condition (**SMC**):
  $\forall x\ \psi(x) \rightarrow cpt(x)$; $\psi$: arbitrary concept

- Necessary and Sufficient Membership Condition (**NSMC**):
  $\forall x\ cpt(x) \leftrightarrow \psi(x)$; $\psi$: arbitrary concept

  o Subsumption Link with Differentia (**SLD**):
     $\forall x\ cpt(x) \rightarrow cpt'(x) \wedge \delta(x)$; cpt': concept of the ontology; $\delta$: arbitrary concept

- Necessary and Sufficient Identity Condition (**NSIC**):
  $\forall x,y\ (cpt(x) \wedge cpt(y) \rightarrow (rel(x,y) \leftrightarrow x=y))$; rel: arbitrary binary relation

- Necessary Identity Condition (**NIC**):
  $\forall x,y\ (cpt(x) \wedge cpt(y) \rightarrow (rel(x,y) \leftarrow x=y))$; rel: arbitrary binary relation

- Sufficient Identity Condition (**SIC**):
  $\forall x,y\ (cpt(x) \wedge cpt(y) \rightarrow (rel(x,y) \rightarrow x=y))$; rel: arbitrary binary relation

- Unity Condition (**UC**):
  $\forall x,t\ cpt(x) \rightarrow [(ED(x) \rightarrow (\forall y,z(P(y,x,t) \wedge P(z,x,t)) \rightarrow rel(y,z)$
  $\wedge \forall y,z(\neg P(y,x,t) \wedge \neg P(z,x,t)) \rightarrow \neg rel(y,z)))$
  $\wedge ((PD(x) \vee AB(x)) \rightarrow (\forall y,z(P(y,x) \wedge P(z,x)) \rightarrow rel(y,z)$
  $\wedge \forall y,z(\neg P(y,x) \wedge \neg P(z,x)) \rightarrow \neg rel(y,z)))]$ ;
  rel: arbitrary binary relation;

## A1.2 Conditions characterizing relations

rel: n-ary relation of the ontology

- Necessary Membership Condition (**NMC**):
  $\forall x_1,x_2\ldots,x_n\ rel(x_1,x_2\ldots,x_n) \rightarrow \psi(x_1,x_2\ldots,x_n)$; $\psi$: arbitrary n-ary relation

    o Subsumption Link (**SL**):
      $\forall x_1,x_2\ldots,x_n\ rel(x_1,x_2\ldots,x_n) \rightarrow rel'(x_1,x_2\ldots,x_n)$; rel': n-ary relation of the ontology

    o Domain Restriction (**DR**):
      $\forall x,y\ rel(x,y) \rightarrow \psi(x)$; rel: binary relation of the ontology; $\psi$: concept of the ontology

    o Disjunctive Domain Restriction (**DDR**):
      $\forall x,y\ rel(x,y) \rightarrow \psi_1(x) \vee \psi_2(x) \ldots \vee \psi_n(x)$; rel: binary relation of the ontology; $\psi_1,\psi_2\ldots,\psi_n$: concepts of the ontology

    o Conjunctive Domain Restriction (**CDR**):
      $\forall x,y\ rel(x,y) \rightarrow \psi_1(x) \wedge \psi_2(x) \ldots \wedge \psi_n(x)$; rel: binary relation of the ontology; $\psi_1,\psi_2\ldots,\psi_n$: concepts of the ontology

    o Range Restriction (**RR**):
      $\forall x,y\ rel(x,y) \rightarrow \psi(y)$; rel: binary relation of the ontology; $\psi$: arbitrary concept

    o Disjunctive Range Restriction (**DRR**):
      $\forall x,y\ rel(x,y) \rightarrow \psi_1(y) \vee \psi_2(y) \ldots \vee \psi_n(y)$; rel: binary relation of the ontology; $\psi_1,\psi_2\ldots,\psi_n$: concepts of the ontology

    o Conjunctive Range Restriction (**CRR**):
      $\forall x,y\ rel(x,y) \rightarrow \psi_1(y) \wedge \psi_2(y) \ldots \wedge \psi_n(y)$; rel: binary relation of the ontology; $\psi_1,\psi_2\ldots,\psi_n$: concepts of the ontology

- o Value Arguments Restriction (**VR1 & VR2 … & VRn**) :
  $\forall x_1, x_2 \ldots, x_n \; rel(x_1, x_2 \ldots, x_n) \rightarrow \psi_1(x_1) \wedge \psi_2(x_2) \ldots \wedge \psi_n(x_n)$ ; $\psi_1, \psi_2 \ldots \psi_n$ : arbitrary concepts

- o Incompatibility Link (IL):
  $\forall x_1, x_2 \ldots, x_n \; rel(x_1, x_2 \ldots, x_n) \rightarrow \neg rel'(x_1, x_2 \ldots, x_n)$; rel': n-ary relation of the ontology

- Sufficient Membership Condition (**SMC**):
  $\forall x_1, x_2 \ldots, x_n \; \psi(x_1, x_2 \ldots, x_n) \rightarrow rel(x_1, x_2 \ldots, x_n)$; $\psi$: arbitrary n-ary relation

- Necessary and Sufficient Membership Condition (**NSMC**):
  $\forall x_1, x_2 \ldots, x_n \; rel(x_1, x_2 \ldots, x_n) \leftrightarrow \psi(x_1, x_2 \ldots, x_n)$; $\psi$: arbitrary n-ary relation

  - o Subsumption Link with Differentia (**SLD**):
    $\forall x_1, x_2 \ldots, x_n \; rel(x_1, x_2 \ldots, x_n) \leftrightarrow rel'(x_1, x_2 \ldots, x_n) \wedge \delta(x_1, x_2 \ldots, x_n)$; rel': n-ary relation of the ontology; $\delta$: arbitrary n-ary relation

  - o Inverse Link (**IVL**):
    $\forall x, y \; rel(x, y) \leftrightarrow rel'(y, x)$; rel, rel': binary relations of the ontology

# Appendix 2: DOLCE-OS

## A2.1 Rigid concepts

Particular, **PT**
**Meta-properties**
PARTICULAR is RIGID (+**R**). PARTICULAR is NOT CARRYING AN IDENTITY CRITERION (-**I**). PARTICULAR is NOT CARRYING A COMMON UNITY CRITERION (-**U**). PARTICULAR is NON-EXTERNALLY-DEPENDENT (-**D**). PARTICULAR is NON-EMPTY. ABSTRACT, ENDURANT, PERDURANT and QUALITY *is a non-trivial Partition of* PARTICULAR.

Abstract, **AB**
**Meta-properties**
ABSTRACT is RIGID (+**R**). ABSTRACT is NOT CARRYING AN IDENTITY CRITERION (-**I**). ABSTRACT is NOT CARRYING A COMMON UNITY CRITERION (-**U**). ABSTRACT is NON-EXTERNALLY-DEPENDENT (-**D**). ABSTRACT is NON-EMPTY.
**Properties**
[EP/SL] An ABSTRACT is a PARTICULAR. [**Ad3a'**; EP/VR] An ABSTRACT *has for parts* only ABSTRACTS. [**Ad3b**; EP/VR] An ABSTRACT *is a part of* only ABSTRACTS. [**Ad5a**; EP/NMC] Every ABSTRACT *is part of* itself. [**Ad8a'**; EP/NMC] Every ABSTRACT1 which *is not a part of* an ABSTRACT2 is such that there exists at least one ABSTRACT3 which *is a part of* ABSTRACT1 and which does not *overlap with* ABSTRACT2.
**Comment**
[CIT] [D18, p. 10] "Abstracts possess no causal power while concretes do." [CIT] [D18, p. 18] "The main characteristic of abstract entities is that they do not have spatial nor temporal qualities, and they are not qualities themselves."

Region, **R**
**Meta-properties**
REGION is RIGID (+**R**). REGION is SUPPLYING AN IDENTITY CRITERION (+**O**). REGION has ANTI-UNITY (~**U**). REGION is NON-EXTERNALLY-DEPENDENT (-**D**). REGION is NON-EMPTY. ABSTRACT REGION, PHYSICAL REGION and TEMPORAL REGION *is a non-trivial Partition of* REGION.
**Properties**
[EP/SL] A REGION is an ABSTRACT. [EP/NSIC] Two REGIONS are the same iff they have the same parts.

Abstract region, **AR**
**Meta-properties**
ABSTRACT REGION is RIGID (+**R**). ABSTRACT REGION is CARRYING AN IDENTITY CRITERION (+**I**). ABSTRACT REGION has ANTI-UNITY (~**U**). ABSTRACT REGION is NON-EXTERNALLY-DEPENDENT (-**D**). ABSTRACT REGION is NON-EMPTY.
**Properties**

[EP/SL] An ABSTRACT REGION is a REGION. [**Ad60b'**; EP/VR] An ABSTRACT REGION *is the quale of* only ABSTRACT QUALITIES *during* a TIME INTERVAL.
**Comment**
[EX] An example of ABSTRACT REGION is the (conventional) value of 1 Euro.

### Physical region, PR
**Meta-properties**
PHYSICAL REGION is RIGID (+**R**). PHYSICAL REGION is CARRYING AN IDENTITY CRITERION (+**I**). PHYSICAL REGION has ANTI-UNITY (~**U**). PHYSICAL REGION is NON-EXTERNALLY-DEPENDENT (-**D**). PHYSICAL REGION is NON-EMPTY.
**Properties**
[EP/SL] A PHYSICAL REGION is a REGION. [**Ad59b'**; EP/VR] A PHYSICAL REGION *is the quale of* only PHYSICAL QUALITIES *during* a TIME INTERVAL
**Comment**
[EX] Examples of PHYSICAL REGIONS are the physical space, an area in the colour spectrum, 80kg.

### Space Region, S
**Meta-properties**
SPACE REGION is RIGID (+**R**). SPACE REGION is CARRYING AN IDENTITY CRITERION (+**I**). SPACE REGION has ANTI-UNITY (~**U**). SPACE REGION is NON-EXTERNALLY-DEPENDENT (-**D**). SPACE REGION is NON-EMPTY.
**Properties**
|EP/SL] A SPACE REGION is a PHYSICAL REGION.

### Temporal region, TR
**Meta-properties**
TEMPORAL REGION is RIGID (+**R**). TEMPORAL REGION is CARRYING AN IDENTITY CRITERION (+**I**). TEMPORAL REGION has ANTI-UNITY (~**U**). TEMPORAL REGION is NON-EXTERNALLY-DEPENDENT (-**D**). TEMPORAL REGION is NON-EMPTY.
**Properties**
[EP/SL] A TEMPORAL REGION is a REGION.
**Comment**
[EX] Examples of TEMPORAL REGIONS are the time axis, 22 June 2002, one second.

### Time interval, T
**Meta-properties**
TIME INTERVAL is RIGID (+**R**). TIME INTERVAL is CARRYING AN IDENTITY CRITERION (+**I**). TIME INTERVAL has ANTI-UNITY (~**U**). TIME INTERVAL is NON-EXTERNALLY-DEPENDENT (-**D**). TIME INTERVAL is NON-EMPTY.
**Properties**
[EP/SL] A TIME INTERVAL is a TEMPORAL REGION.

### Endurant, continuant, ED
**Meta-properties**
ENDURANT is RIGID (+**R**). ENDURANT is NOT CARRYING AN IDENTITY CRITERION (-**I**). ENDURANT is NOT CARRYING A COMMON UNITY CRITERION

(**-U**). ENDURANT is NON-EXTERNALLY-DEPENDENT (**-D**). ENDURANT *partially generically spatially depends on* PERDURANT. ENDURANT is NON-EMPTY. ARBITRARY SUM, NON-PHYSICAL ENDURANT and PHYSICAL ENDURANT *is a non-trivial Partition of* ENDURANT.

**Properties**
[EP/SL] An ENDURANT, or "CONTINUANT", is a PARTICULAR. [**Td15a'**; EP/ER] Every ENDURANT *is present at* at least one TIME INTERVAL. [**Ad14'**; EP/NMC] For every ENDURANT1 which *is present at* the same TIME INTERVAL of an ENDURANT2 and which *is not a part of* the ENDURANT2 *during* this TIME INTERVAL, there exists an ENDURANT3 such that the ENDURANT3 *is a part of* the ENDURANT1 *at* a TIME INTERVAL and the ENDURANT3 does not *overlap with* the ENDURANT2 *at* that TIME INTERVAL. [**Ad16'**; EP/NMC] Every ENDURANT which *is present at* a TIME INTERVAL *is a part of* itself *during* that TIME INTERVAL. [**Ad35'**; EP/ER] Every ENDURANT *participates in* at least one PERDURANT *during* at least one TIME INTERVAL. [**Td1a'**; EP/NMC] No ENDURANT *constitutes* itself *during* a TIME INTERVAL. [**Td6'**; EP/NMC] No ENDURANT *participates in* itself *during* a TIME INTERVAL.

**Comment**
[SA] ENDURANTS are divided into PHYSICAL ENDURANTS and NON-PHYSICAL ENDURANTS according to whether or not they have direct spatial qualities. [CIT] [D18, p. 15] "Endurants are *wholly* present (i.e., all their proper parts are present) at any time they are present." [CIT] [D18, p. 16] "Endurants can "genuinely" change in time, in the sense that the very same endurant as a whole can have incompatible properties at different times."

Arbitrary sum, **AS**
**Meta-properties**
ARBITRARY SUM is RIGID (+**R**). ARBITRARY SUM is SUPPLYING AN IDENTITY CRITERION (+**O**). ARBITRARY SUM has ANTI-UNITY (**~U**). ARBITRARY SUM is NON-EXTERNALLY-DEPENDENT (**-D**). ARBITRARY SUM is NON-EMPTY.

**Properties**
[EP/SL] An ARBITRARY SUM is an ENDURANT. [EP/NSIC] Two ARBITRARY SUMS are the same iff they are the sum of the same entities.

**Comment**
[EX] An example of ARBITRARY SUM is a left foot plus a car.

Non-Physical endurant, **NPED**
**Meta-properties**
NON-PHYSICAL ENDURANT is RIGID (+**R**). NON-PHYSICAL ENDURANT is NOT CARRYING AN IDENTITY CRITERION (**-I**). NON-PHYSICAL ENDURANT is NOT CARRYING A COMMON UNITY CRITERION (**-U**). NON-PHYSICAL ENDURANT is EXTERNALLY-DEPENDENT (+**D**). NON-PHYSICAL ENDURANT *mutually specifically constantly depends on* ABSTRACT QUALITY. [**Ad74**] NON-PHYSICAL ENDURANT *one-sided constantly depends on* PHYSICAL ENDURANT. NON-PHYSICAL ENDURANT is NON-EMPTY.

**Properties**
[EP/SL] A NON-PHYSICAL ENDURANT is an ENDURANT. [**Ad12a'**; EP/VR] A NON-PHYSICAL ENDURANT *has for parts* only NON-PHYSICAL ENDURANTS *during* a TIME INTERVAL. [**Ad12b'**; EP/VR] A NON-PHYSICAL ENDURANT *is part of* only NON-PHYSICAL ENDURANTS *during* a TIME INTERVAL. [**Ad22a'**; EP/VR] A NON-PHYSICAL ENDURANT *has for constituents* only NON-PHYSICAL

ENDURANTS *during* a TIME INTERVAL. [**Ad22b'**; EP/VR] A NON-PHYSICAL ENDURANT *constitutes* only NON-PHYSICAL ENDURANTS *during* a TIME INTERVAL. [**Ad41ab'**; EP/VR] A NON-PHYSICAL ENDURANT *has for qualities* only ABSTRACT QUALITIES.

Non-Physical object, **NPOB**
**Meta-properties**
NON-PHYSICAL OBJECT is RIGID (+**R**). NON-PHYSICAL OBJECT is NOT CARRYING AN IDENTITY CRITERION (-**I**). NON-PHYSICAL OBJECT is NOT CARRYING A COMMON UNITY CRITERION (-**U**). NON-PHYSICAL OBJECT is EXTERNALLY-DEPENDENT (+**D**). NON-PHYSICAL OBJECT is NON-EMPTY. MENTAL OBJECT and SOCIAL OBJECT *is a non-trivial Partition of* NON-PHYSICAL OBJECT.
**Properties**
[EP/SL] A NON-PHYSICAL OBJECT is a NON-PHYSICAL ENDURANT.
**Comment**
[SA] NON-PHYSICAL OBJECTS are divided into SOCIAL OBJECTS and MENTAL OBJECTS according to whether or not they are generically dependent on a community of agents.

Mental object, **MOB**
**Meta-properties**
MENTAL OBJECT is RIGID (+**R**). MENTAL OBJECT is NOT CARRYING AN IDENTITY CRITERION (-**I**). MENTAL OBJECT is NOT CARRYING A COMMON UNITY CRITERION (-**U**). MENTAL OBJECT is EXTERNALLY-DEPENDENT (+**D**). [**Ad71**] MENTAL OBJECT *one-sided specifically constantly depends on* AGENTIVE PHYSICAL OBJECT. MENTAL OBJECT is NON-EMPTY.
**Properties**
[EP/SL] A MENTAL OBJECT is a NON-PHYSICAL OBJECT.
**Comment**
[EX] Examples of MENTAL OBJECTS are a percept, a sense datum.

Social object, **SOB**
**Meta-properties**
SOCIAL OBJECT is RIGID (+**R**). SOCIAL OBJECT is NOT CARRYING AN IDENTITY CRITERION (-**I**). SOCIAL OBJECT is NOT CARRYING A COMMON UNITY CRITERION (-**U**). SOCIAL OBJECT is EXTERNALLY-DEPENDENT (+**D**). SOCIAL OBJECT is NON-EMPTY. AGENTIVE SOCIAL OBJECT and NON-AGENTIVE SOCIAL OBJECT *is a non-trivial Partition of* SOCIAL OBJECT.
**Properties**
[EP/SL] A SOCIAL OBJECT is a NON-PHYSICAL OBJECT.
**Comment**
[SA] SOCIAL OBJECTS are divided into AGENTIVE SOCIAL OBJECTS and NON-AGENTIVE SOCIAL OBJECTS whether or not they have intentions, beliefs and desires.

Agentive social object, **ASO**
**Meta-properties**
AGENTIVE SOCIAL OBJECT is RIGID (+**R**). AGENTIVE SOCIAL OBJECT is NOT CARRYING AN IDENTITY CRITERION (-**I**). AGENTIVE SOCIAL OBJECT

is NOT CARRYING A COMMON UNITY CRITERION (**-U**). AGENTIVE SOCIAL OBJECT is EXTERNALLY-DEPENDENT (**+D**). AGENTIVE SOCIAL OBJECT is NON-EMPTY. SOCIAL AGENT and SOCIETY *is a non-trivial Partition of* AGENTIVE SOCIAL OBJECT.
**Properties**
[EP/SL] An AGENTIVE SOCIAL OBJECT is a SOCIAL OBJECT.

### Social agent, SAG
**Meta-properties**
SOCIAL AGENT is RIGID (**+R**). SOCIAL AGENT is NOT CARRYING AN IDENTITY CRITERION (**-I**). SOCIAL AGENT is NOT CARRYING A COMMON UNITY CRITERION (**-U**). SOCIAL AGENT is EXTERNALLY-DEPENDENT (**+D**). [**Ad72**] SOCIAL AGENT *one-sided generically constantly depends on* AGENTIVE PHYSICAL OBJECT. SOCIAL AGENT is NON-EMPTY.
**Properties**
[EP/SL] A SOCIAL AGENT is an AGENTIVE SOCIAL OBJECT.
**Comment**
[EX] Examples of SOCIAL AGENTS are a (legal) person, a contractant.

### Society, SC
**Meta-properties**
SOCIETY is RIGID (**+R**). SOCIETY is NOT CARRYING AN IDENTITY CRITERION (**-I**). SOCIETY is CARRYING A COMMON UNITY CRITERION (**+U**). SOCIETY is EXTERNALLY-DEPENDENT (**+D**). [**Ad32**] SOCIETY *is constantly generically constituted by* SOCIAL AGENT. SOCIETY is NON-EMPTY.
**Properties**
[EP/SL] A SOCIETY is an AGENTIVE SOCIAL OBJECT.
**Comment**
[EX] Examples of SOCIETIES are Fiat, Apple, the Bank of Italy.

### Non-agentive social object, NASO
**Meta-properties**
NON-AGENTIVE SOCIAL OBJECT is RIGID (**+R**). NON-AGENTIVE SOCIAL OBJECT is NOT CARRYING AN IDENTITY CRITERION (**-I**). NON-AGENTIVE SOCIAL AGENT is NOT CARRYING A COMMON UNITY CRITERION (**-U**). NON-AGENTIVE SOCIAL OBJECT is EXTERNALLY-DEPENDENT (**+D**). [**Ad73**] NON-AGENTIVE SOCIAL OBJECT *one-sided generically constantly depends on* SOCIETY. NON-AGENTIVE SOCIAL OBJECT is NON-EMPTY.
**Properties**
[EP/SL] A NON-AGENTIVE SOCIAL OBJECT is a SOCIAL OBJECT.
**Comment**
[EX] Examples of NON-AGENTIVE SOCIAL OBJECTS are a law, an economic system, a currency, an asset.

### Physical endurant, PED
**Meta-properties**
PHYSICAL ENDURANT is RIGID (**+R**). PHYSICAL ENDURANT is NOT CARRYING AN IDENTITY CRITERION (**-I**). PHYSICAL ENDURANT is NOT CARRYING A COMMON UNITY CRITERION (**-U**). PHYSICAL ENDURANT is NON-EXTERNALLY-DEPENDENT (**-D**). PHYSICAL ENDURANT *mutually* *specifically spatially depends on* PHYSICAL QUALITY. PHYSICAL ENDURANT is NON-EMPTY. AMOUNT OF MATTER, FEATURE and PHYSICAL OBJECT *is a non-trivial Partition of* PHYSICAL ENDURANT.

**Properties**
[EP/SL] A PHYSICAL ENDURANT is an ENDURANT. [**Ad11a'**; EP/VR] A PHYSICAL ENDURANT *has for parts* only PHYSICAL ENDURANTS *during* a TIME INTERVAL. [**Ad11b'**; EP/VR] A PHYSICAL ENDURANT *is part of* only PHYSICAL ENDURANTS *during* a TIME INTERVAL. [**Ad19'**; EP/NMC] Every PHYSICAL ENDURANT which *is a part of* another PHYSICAL ENDURANT *at* a TIME INTERVAL *is spatially included in* this other PHYSICAL ENDURANT *during* that TIME INTERVAL. [**Ad21a'**; EP/VR] A PHYSICAL ENDURANT *has for constituents* only PHYSICAL ENDURANT *during* a TIME INTERVAL. [**Ad21b'**; EP/VR] A PHYSICAL ENDURANT *constitutes* only PHYSICAL ENDURANTS *during* a TIME INTERVAL. [**Ad28'**; EP/NMC] Every PHYSICAL ENDURANT which *constitutes* another PHYSICAL ENDURANT *during* a TIME INTERVAL *temporarily spatially coincides with* this other PHYSICAL ENDURANT *during* the TIME INTERVAL. [**Ad40ab'**; EP/VR] A PHYSICAL ENDURANT *has for qualities* only PHYSICAL QUALITIES. [**Ad50'**; EP/ER] Every PHYSICAL ENDURANT *has for quality* at least one SPATIAL LOCATION. [**Td16a'**; EP/NMC] Every PHYSICAL ENDURANT which *is present at* a TIME INTERVAL *is present in* at least one SPACE REGION *at* that TIME INTERVAL.

**Comment**
[CIT] [D18, p. 22] "Within physical endurants, we distinguish between *amounts of matter*, *objects*, and *features*. This distinction is mainly based on the notion of unity we have discussed and formalized in [Gangemi *et al.* 2001]."

Amount of matter, **M**
**Meta-properties**
AMOUNT OF MATTER is RIGID (+**R**). AMOUNT OF MATTER is SUPPLYING AN IDENTITY CRITERION (+**O**). AMOUNT OF MATTER has ANTI-UNITY (~**U**). AMOUNT OF MATTER is NON-EXTERNALLY-DEPENDENT (-**D**). AMOUNT OF MATTER is NON-EMPTY.

**Properties**
[EP/SL] An AMOUNT OF MATTER is a PHYSICAL ENDURANT. [EP/NSIC] Two AMOUNTS OF MATTER are the same iff they have the same parts.

**Comment**
[CIT] [D18, p. 23] "The common trait of *amounts of matter* is that they are endurants with no unity (according to [Gangemi *et al.*, 2001], none of them is an essential whole). Amounts of matter – "stuffs" referred to by mass nouns like "gold", "iron", "wood", "sand", "meat", etc. – are mereologically invariant, in the sense that they change their identity when they change some parts." [EX] Examples of AMOUNTS OF MATTER are some air, some gold, some cement.

Feature, **F**
**Meta-properties**
FEATURE is RIGID (+**R**). FEATURE is NOT CARRYING AN IDENTITY CRITERION (-**I**). FEATURE is NOT CARRYING A COMMON UNITY CRITERION (-**U**). FEATURE is EXTERNALLY-DEPENDENT (+**D**). [**Ad70**] FEATURE *one-sided generically constantly depends on* NON-AGENTIVE PHYSICAL OBJECT. FEATURE is NON-EMPTY.

**Properties**

[EP/SL] A FEATURE is a PHYSICAL ENDURANT.
**Comment**
[CIT] [D18, p. 23] "Typical examples of features are "parasitic entities" such as holes, boundaries, surfaces, or stains, which are generically constantly dependent on physical objects (their host). All features are essential holes, but, as in the case of objects, no common unity criterion may exist for all of them. However, typical features have a topological unity, as they are singular entities. Some features may be *relevant parts* of their host, like a bump or an edge, or *places* like a hole in a piece of cheese, the underneath of a table, the front of a house, which are not parts of their host.". [CIT] [D18, p. 23] "It may be interesting to note that we do not consider body parts like heads or hands as features: the reason is that we assume that a hand can be detached from its host (differently from a hole or a bump), and we assume that in this case it retains its identity. Should we reject this assumption, then body parts would be features.". [EX] Examples of FEATURES are a hole, a gulf, an opening, a boundary.

Physical object, object, **POB**
**Meta-properties**
PHYSICAL OBJECT is RIGID (+**R**). PHYSICAL OBJECT is SUPPLYING AN IDENTITY CRITERION (+**O**). PHYSICAL OBJECT is not CARRYING A COMMON UNITY CRITERION (-**U**). PHYSICAL OBJECT is NON-EXTERNALLY-DEPENDENT (-**D**). PHYSICAL OBJECT is NON-EMPTY. AGENTIVE PHYSICAL OBJECT and NON-AGENTIVE PHYSICAL OBJECT *is a non-trivial Partition of* PHYSICAL OBJECT.
**Properties**
[EP/SL] A PHYSICAL OBJECT, or "OBJECT", is a PHYSICAL ENDURANT.
[EP/NSIC] Two PHYSICAL OBJECTS are the same iff they have the same spatial location at the same time.
**Comment**
[CIT] [D18, p. 23] "The main characteristic of objects is that they are endurants with unity. However, they have no *common* unity criterion, since different subtypes of objects may have different unity criteria. Differently from aggregates, (most) objects change some of their parts while keeping their identity, they can have therefore *temporary parts*."

Agentive physical object, **APO**
**Meta-properties**
AGENTIVE PHYSICAL OBJECT is RIGID (+**R**). AGENTIVE PHYSICAL OBJECT is CARRYING AN IDENTITY CRITERION (+**I**). AGENTIVE PHYSICAL OBJECT is not CARRYING A COMMON UNITY CRITERION (-**U**). AGENTIVE PHYSICAL OBJECT is NON-EXTERNALLY-DEPENDENT (-**D**). [**Ad31**] AGENTIVE PHYSICAL OBJECT *is constantly generically constituted by* NON-AGENTIVE PHYSICAL OBJECT. AGENTIVE PHYSICAL OBJECT is NON-EMPTY.
**Properties**
[PE/SL] An AGENTIVE PHYSICAL OBJECT is a PHYSICAL OBJECT.
**Comment**
[CIT] [D18, p. 23] "Within physical objects, a special place have those to which we ascribe *intentions*, *beliefs*, and *desires*. These are called *Agentive*, as opposite to *Non-agentive*. Intentionality is understood as the capability of heading for/dealing with objects or states of the world… In general, we assume that agentive objects are *constituted* by no-agentive objects: a person is constituted by an organism, a robot is

constituted by some machinery, and so on." [EX] A human person (as opposed to legal person) is an example of AGENTIVE PHYSICAL OBJECT.

Non-agentive physical object, **NAPO**
**Meta-properties**
NON-AGENTIVE PHYSICAL OBJECT is RIGID (+**R**). NON-AGENTIVE PHYSICAL OBJECT is CARRYING AN IDENTITY CRITERION (+**I**). NON-AGENTIVE PHYSICAL OBJECT is not CARRYING A COMMON UNITY CRITERION (-**U**). NON-AGENTIVE PHYSICAL OBJECT is NON-EXTERNALLY-DEPENDENT (-**D**). [**Ad30**] NON-AGENTIVE PHYSICAL OBJECT *is constantly generically constituted by* AMOUNT OF MATTER. NON-AGENTIVE PHYSICAL OBJECT is NON-EMPTY.
**Properties**
[EP/SL] A NON-AGENTIVE PHYSICAL OBJECT is a PHYSICAL OBJECT.
**Comment**
[EX] Examples of NON-AGENTIVE PHYSICAL OBJECTS are a hammer, a house, an opening, a boundary.

Perdurant, occurrent, **PD**
**Meta-properties**
PERDURANT is RIGID (+**R**). PERDURANT is NOT CARRYING AN IDENTITY CRITERION (-**I**). PERDURANT is NOT CARRYING A COMMON UNITY CRITERION (-**U**). PERDURANT is EXTERNALLY-DEPENDENT (+**D**). PERDURANT *mutually specifically constantly depends on* TEMPORAL QUALITY. PERDURANT *inversely partially generically spatially depends on* ENDURANT. PERDURANT is NON-EMPTY. EVENT and STATIVE *is a non-trivial Partition of* PERDURANT.
**Properties**
[EP/SL] A PERDURANT, or "OCCURRENT", is a PARTICULAR. [**Ad2a'**; EP/VR] A PERDURANT *has for parts* only PERDURANTS. [**Ad2b**; EP/VR] A PERDURANT *is a part of* only PERDURANTS. [**Ad5b**; EP/NMC] Every PERDURANT *is part of* itself. [**Td15b'**; EP/ER] Every PERDURANT *is present at* at least one TIME INTERVAL. [**Ad8b'**; EP/NMC] Every PERDURANT1 which *is not a part of* a PERDURANT2 is such that there exists at least one PERDURANT3 which *is a part of* PERDURANT1 and which does not *overlap with* PERDURANT2. [**Ad23a'**; EP/VR] A PERDURANT *has for constituents* only PERDURANTS *during* a TIME INTERVAL. [**Ad23b'**; EP/VR] A PERDURANT *constitutes* only PERDURANTS *during* a TIME INTERVAL. [**Ad34'**; NMC] For every PERDURANT *present at* a TIME INTERVAL there exists at least one ENDURANT which *participates in* the PERDURANT *during* that TIME INTERVAL. [**Ad39ab'**; EP/VR] A PERDURANT *has for qualities* only TEMPORAL QUALITIES. [**Ad49'**; EP/ER] Every PERDURANT *has for quality* at least one TEMPORAL LOCATION. [**Td1b'**; EP/NMC] No PERDURANT *constitutes* itself *during* a TIME INTERVAL.
**Comment**
[SA] PERDURANTS are divided among STATIVES and EVENTS according to whether they hold of the mereological sum of two of their instances, *i.e.* if they are cumulative or not. [CIT] [D18, p. 15] "Perdurants […] just extend in time by accumulating different temporal parts, so that, at any time they are present, they are only *partially* present, in the sense that some of their proper temporal parts (e.g., their previous or future phases) may be not present." [CIT] [D18, p. 16] "Perdurants cannot change […] since none of their parts keeps its identity in time." [CIT] [D18, p.24] "They can have temporal parts or spatial parts. For

instance, the first movement of (an execution of) a symphony is a temporal part of it. On the other side, the play performed by the left side of the orchestra is a spatial part. In both cases, these parts are occurrences themselves."

Event, **EV**
**Meta-properties**
EVENT is RIGID (+**R**). EVENT is NOT CARRYING AN IDENTITY CRITERION (-**I**). EVENT is NOT CARRYING A COMMON UNITY CRITERION (-**U**). EVENT is EXTERNALLY-DEPENDENT (+**D**). EVENT is ANTI-CUMULATIVE. EVENT is NON-EMPTY. ACCOMPLISHMENT and ACHIEVEMENT *is a non-trivial Partition of* EVENT.
**Properties**
[EP/SL] An EVENT is a PERDURANT.
**Comment**
[SA] EVENTS are divided among ACHIEVEMENTS and ACCOMPLISHMENTS whether they are atomic or not.

Accomplishment, **ACC**
**Meta-properties**
ACCOMPLISHMENT is RIGID (+**R**). ACCOMPLISHMENT is NOT CARRYING AN IDENTITY CRITERION (-**I**). ACCOMPLISHMENT is NOT CARRYING A COMMON UNITY CRITERION (-**U**). ACCOMPLISHMENT is EXTERNALLY-DEPENDENT (+**D**). ACCOMPLISHMENT is ANTI-CUMULATIVE. ACCOMPLISHMENT is ANTI-ATOMIC. ACCOMPLISHMENT is NON-EMPTY.
**Properties**
[EP/SLD] An ACCOMPLISHMENT is an EVENT which is not ATOMIC.
**Comment**
[EX] Examples of ACCOMPLISHMENTS are a conference, an ascent, a performance.

Achievement, **ACH**
**Meta-properties**
ACHIEVEMENT is RIGID (+**R**). ACHIEVEMENT is NOT CARRYING AN IDENTITY CRITERION (-**I**). ACHIEVEMENT is NOT CARRYING A COMMON UNITY CRITERION (-**U**). ACHIEVEMENT is EXTERNALLY-DEPENDENT (+**D**). ACHIEVEMENT is ANTI-CUMULATIVE. ACHIEVEMENT is ATOMIC. ACHIEVEMENT is NON-EMPTY.
**Properties**
[EP/SLD] An ACHIEVEMENT is an EVENT which is ATOMIC.
**Comment**
[EX] Examples of ACHIEVEMENTS are reaching the summit of K2, a departure, a death.

Stative, **STV**
**Meta-properties**
STATIVE is RIGID (+**R**). STATIVE is NOT CARRYING AN IDENTITY CRITERION (-**I**). STATIVE is NOT CARRYING A COMMON UNITY CRITERION (-**U**). STATIVE is EXTERNALLY-DEPENDENT (+**D**). STATIVE is NON-EMPTY. STATIVE is CUMULATIVE. PROCESS and STATE *is a non-trivial Partition of* STATIVE.
**Properties**
[EP/SL] A STATIVE is a PERDURANT.

**Comment**
[EX] A sitting is STATIVE since the sum of two sittings is still a sitting. [SA] STATIVES are divided among STATES and PROCESSES according to homeomericity.

## Process, PRO
**Meta-properties**
PROCESS is RIGID (+**R**). PROCESS is NOT CARRYING AN IDENTITY CRITERION (-**I**). PROCESS is NOT CARRYING A COMMON UNITY CRITERION (-**U**). PROCESS is EXTERNALLY-DEPENDENT (+**D**). PROCESS is CUMULATIVE. PROCESS is ANTI-HOMEOMEROUS. PROCESS is NON-EMPTY.
**Properties**
[EP/SL] A PROCESS is a STATIVE.
**Comment**
[CIT] [D18, p. 24] "*running* is classified as a process since there are (very short) temporal parts of a running that are not themselves runnings." [EX] Examples of PROCESSES are running, writing.

## State, ST
**Meta-properties**
STATE is RIGID (+**R**). STATE is NOT CARRYING AN IDENTITY CRITERION (-**I**). STATE is NOT CARRYING A COMMON UNITY CRITERION (-**U**). STATE is EXTERNALLY-DEPENDENT (+**D**). STATE is CUMULATIVE. STATE is HOMEOMEROUS. STATE is NON-EMPTY.
**Properties**
[EP/SL] A STATE is a STATIVE.
**Comment**
[EX] Examples of STATES are being sitting, being open, being happy, being red.

## Quality, Q
**Meta-properties**
QUALITY is RIGID (+**R**). QUALITY is NOT CARRYING AN IDENTITY CRITERION (-**I**). QUALITY is EXTERNALLY-DEPENDENT (+**D**). QUALITY is NON-EMPTY. ABSTRACT QUALITY, PHYSICAL QUALITY and TEMPORAL QUALITY *is a non-trivial Partition of* QUALITY.
**Properties**
[EP/SL] A QUALITY is a PARTICULAR. [**Td15c'**; EP/ER] Every QUALITY *is present at* at least one TIME INTERVAL. [**Td8'**; EP/NMC] No QUALITY *is a quality of* itself.
**Comment**
[CIT] [D18, p. 16] "Qualities can be seen as the basic entities we can perceive or measure: shapes, colours, sizes, sounds, as well as weights, lengths, electrical charges." [CIT] [D18, p. 16] "Qualities *inhere* to entities: every entity (including qualities themselves) comes with certain qualities, which exist as long as the entity exists." [CIT] [D18, p. 16] "No two particulars can have the same quality, and each quality is *specifically constantly dependent* on the entity it inheres in: at any time, a quality can't be present unless the entity it inheres in is also present." [CIT] [D18, p.17] "Each quality type has an associated quality space with a specific structure. For example, lengths are usually associated to a metric linear space, and colours to a topological 2D space." [CIT] [D18, p. 18] "Since no parthood is defined, qualities are neither endurants nor perdurants, although their persistence conditions may be similar, in certain cases, to those of endurants or perdurants."

Abstract quality, **AQ**
**Meta-properties**
  ABSTRACT QUALITY is RIGID (+**R**). ABSTRACT QUALITY is NOT CARRYING AN IDENTITY CRITERION (-**I**). ABSTRACT QUALITY is EXTERNALLY-DEPENDENT (+**D**). [**Ad69**] ABSTRACT QUALITY *mutually specifically constantly depends on* NON-PHYSICAL ENDURANT. ABSTRACT QUALITY is NON-EMPTY.
**Properties**
  [EP/SL] An ABSTRACT QUALITY is a QUALITY. [**Ad41aa'**; EP/VR] An ABSTRACT QUALITY *has for qualities* only ABSTRACT QUALITIES. [**Ad41b**; EP/EVR] An ABSTRACT QUALITY *is a quality of* only ABSTRACT QUALITIES or NON-PHYSICAL ENDURANTS. [**Ad48**; EP/ER] Every ABSTRACT QUALITY *is a quality of* exactly one NON-PHYSICAL ENDURANT. [**Ad60a'**; EP/VR] An ABSTRACT QUALITY *has for quales* only ABSTRACT REGIONS *during* a TIME INTERVAL. [**Ad62b'**; EP/NMC] Every ABSTRACT QUALITY which *is present at* a TIME INTERVAL *has for quale* at least one ABSTRACT REGION *during* that TIME INTERVAL.
**Comment**
  [EX] The value of an asset is an example of ABSTRACT QUALITY.

Physical quality, **PQ**
**Meta-properties**
  PHYSICAL QUALITY is RIGID (+**R**). PHYSICAL QUALITY is NOT CARRYING AN IDENTITY CRITERION (-**I**). PHYSICAL QUALITY is EXTERNALLY-DEPENDENT (+**D**). [**Ad68**] PHYSICAL QUALITY *mutually specifically spatially depends on* PHYSICAL ENDURANT. PHYSICAL QUALITY is NON-EMPTY.
**Properties**
  [EP/SLD] A PHYSICAL QUALITY is a QUALITY which directly inheres to PHYSICAL ENDURANTS. [**Ad40aa'**; EP/VR] A PHYSICAL QUALITY *has for qualities* only PHYSICAL QUALITIES. [**Ad40b**; EP/EVR] A PHYSICAL QUALITY *is a quality of* only PHYSICAL QUALITIES or PHYSICAL ENDURANTS. [**Ad47**; EP/ER] Every PHYSICAL QUALITY *is a quality of* exactly one PHYSICAL ENDURANT. [**Ad59a'**; EP/VR] A PHYSICAL QUALITY *has for quales* only PHYSICAL REGIONS *during* a TIME INTERVAL. [**Ad62a'**; EP/NMC] Every PHYSICAL QUALITY which *is present at* a TIME INTERVAL *has for quale* at least one PHYSICAL REGION *during* that TIME INTERVAL. [**Td16b'**; EP/NMC] Every PHYSICAL QUALITY which *is present at* a TIME INTERVAL *is present in* at least one SPACE REGION *at* that TIME INTERVAL.
**Comment**
  [EX] Examples of PHYSICAL QUALITIES are the weight of a pen, the colour of an apple.

Spatial location, **SL**
**Meta-properties**
  SPATIAL LOCATION is RIGID (+**R**). SPATIAL LOCATION is NOT CARRYING AN IDENTITY CRITERION (-**I**). SPATIAL LOCATION is EXTERNALLY-DEPENDENT (+**D**). SPATIAL LOCATION is NON-EMPTY.
**Properties**
  [EP/SL] A SPATIAL LOCATION is a PHYSICAL QUALITY. [**Ad61'**; EP/VR] A SPATIAL LOCATION *has for quale* only SPACE REGIONS *during* a TIME INTERVAL.

Temporal quality, **TQ**
**Meta-properties**
TEMPORAL QUALITY is RIGID (+**R**). TEMPORAL QUALITY is NOT CARRYING AN IDENTITY CRITERION (-**I**). TEMPORAL QUALITY is EXTERNALLY-DEPENDENT (+**D**). [**Ad67**] TEMPORAL QUALITY *mutually specifically constantly depends on* PERDURANT. TEMPORAL QUALITY is NON-EMPTY.
**Properties**
[EP/SLD] A TEMPORAL QUALITY is a QUALITY which directly inheres to PERDURANTS. [**Ad39aa'**; EP/VR] A TEMPORAL QUALITY *has for qualities* only TEMPORAL QUALITIES. [**Ad39b**; EP/EVR] A TEMPORAL QUALITY *is a quality of* only TEMPORAL QUALITIES or PERDURANTS. [**Ad46**; EP/ER] Every TEMPORAL QUALITY *is a quality of* exactly one PERDURANT. [**Ad55'**; EP/ER] Every TEMPORAL QUALITY *has for quale* at least one TEMPORAL REGION.
**Comment**
[EX] Examples of TEMPORAL QUALITIES are the duration of World War I, the starting time of the 2000 Olympics.

Temporal location, **TL**
**Meta-properties**
TEMPORAL LOCATION is RIGID (+**R**). TEMPORAL LOCATION is NOT CARRYING AN IDENTITY CRITERION (-**I**). TEMPORAL LOCATION is EXTERNALLY-DEPENDENT (+**D**). TEMPORAL LOCATION is NON-EMPTY.
**Properties**
[EP/SL] A TEMPORAL LOCATION is a TEMPORAL QUALITY. [**Ad53'**; EP/VR] A TEMPORAL LOCATION *has for quale* only TIME INTERVALS.

### A2.2 Non-rigid concept

Atom, **At**
**Meta-properties**
ATOM is NON-RIGID (-**R**). ATOM is NOT CARRYING AN IDENTITY CRITERION (-**I**). ATOM is NOT CARRYING A COMMON UNITY CRITERION (-**U**). ATOM is NON-EXTERNALLY-DEPENDENT (-**D**).
**Properties**
[**Dd16**; EP/NSMC] x is an ATOM iff there does not exist y such that y *is a proper part of* x.

### A2.3 Binary relations

Depends constantly and specifically on, **SD**
**Properties**
[EP/DDR & DRR] An ENDURANT, a PERDURANT or a QUALITY *depends constantly and specifically on* an ENDURANT, a PERDURANT or a QUALITY. [**Dd69**; EP/NSMC] x *depends constantly and specifically on* y iff necessarily x *is present at* a t and y *is present at* every t such that x *is present at* t.

Depends spatially and specifically on, **SD$_S$**
**Properties**

[**Dd78**; EP/NSMC] x *depends spatially and specifically on* y iff necessarily there exists at least one t and one s such that x *is present in* s *at* t and y *is present in* s *at* t for every s and t such that x *is present in s* at t.

Depends spatially, specifically and partially on, **PSD$_S$**
**Properties**
[**Dd79**; EP/NSMC] x *depends spatially, specifically and partially on* y iff necessarily there exists at least one t and one s such that x *is present in* s *at* t and for every s and t such that x *is present in* s *at* t, there exists at least one s' such that s' *is a proper part of* s and y *is present in* s' *at* t.

Depends spatially, specifically, partially and inversely on, **P$^{-1}$SD$_S$**
**Properties**
[**Dd80**; EP/NSMC] x *depends spatially, specifically, partially and inversely on* y iff necessarily there exists at least one t and one s such that x *is present in* s *at* t and for every s and t such that x *is present in* s *at* t, there exists at least one s' such that s *is a proper part of* s' and y *is present in* s' *at* t.

Has for part, **Pinv**
**Properties**
[EP/DDR & DRR] An ABSTRACT or a PERDURANT *has for part* an ABSTRACT or a PERDURANT. [EP/IVL] *Has for part* mutually implies *is a part of*.

Has for proper part, **PPinv**
**Properties**
[EP/SLD] x *has for proper part* y iff x *has for part* y and not y *has for part* x. [EP/IVL] Has for proper part mutually implies is *a proper part of*.

Has for quale, **qlinv**
**Properties**
[EP/DR & RR] A TEMPORAL QUALITY *has for quale* A TEMPORAL REGION. [EP/IVL] *Has for quale* mutually implies *is the quale of*.

Has for quality, **qtinv**
**Properties**
[EP/DDR & RR] A QUALITY, an ENDURANT or a PERDURANT *has for quality* a QUALITY. [EP/MIL] *Has for quality* mutually implies *is a quality of*.

Has for temporal quale, **ql$_T$inv**
**Properties**
[EP/DDR & RR] A PERDURANT, an ENDURANT or a QUALITY, *has for temporal quale* A TIME INTERVAL. [EP/IVL] *Has for temporal quale* mutually implies *is a temporal quale of*.

Is a constant part, **CP**
**Properties**
[EP/DR & RR] An ENDURANT *is a constant part of* an ENDURANT. [**Dd25**; EP/NSMC] x *is a constant part of* y iff y *is present at* at least one t and x *is a part of* y *during* each t such that y *is present at* that t.

Is a part of, **P**
**Properties**
   [**Ad1**; EP/DDR & DRR] An ABSTRACT or a PERDURANT *is a part of* an ABSTRACT or a PERDURANT. [**Ad6**; EP/NMC] x *is a part of* y implies that if y *is a part of* x then x is equal to y. [**Ad7**; EP/NMC] x *is a part of* y implies that if y *is a part of* z then x *is a part of* z. [EP/IVL] *Is a part of* mutually implies *has for part*.

Is a proper part of, **PP**
**Properties**
   [**Dd14**; EP/SLD] x *is a proper part of* y iff x *is a part of* y and not y *is a part of* x. [EP/IVL] *Is a proper part of* mutually implies *has for proper part*.

Is a spatial part of, **P$_S$**
**Properties**
   [EP/DR & RR] A PERDURANT *is a spatial part of* a PERDURANT. [**Dd55**; EP/SLD] x *is a spatial part of* y iff x *is a part of* y and x is a PERDURANT and x *temporally coincides with* y.

Is a temporal part of, **P$_T$**
**Properties**
   [EP/DR & RR] A PERDURANT *is a temporal part of* a PERDURANT. [**Dd54**; EP/SLD] x *is a temporal part of* y iff x *is a part of* y and x is a PERDURANT and for each z such that z *is a part of* y and z *is temporally included in* x then z *is a part of* x.

Is an atomic part of, **AtP**
**Properties**
   [**Dd17**; EP/SLD] x *is an atomic part of* y iff x *is a part of* y and x is an ATOM.

Is a quality of, **qt**
**Properties**
   [**Ad38**; EP/DR & DRR] A QUALITY *is a quality of* a QUALITY, an ENDURANT or a PERDURANT. [**Ad42**; EP/NMC] x *is a quality of* y implies that if y *is a quality of* z then x *is a quality of* z. [EP/IVL] *Is a quality of* mutually implies *has for quality*.

Is a direct quality of, **dqt**
**Properties**
   [**Dd28**; EP/SLD] x *is a direct quality of* y iff x *is a quality of* y and there does not exist z such that x *is a quality of* z and z *is a quality of* y. [**Ad43**; EP/NC] x *is a direct quality of* y implies that if x *is a direct quality of* y' then y is equal to y'.

Is a temporal quale of, **ql$_T$**
**Properties**
   [EP/DR & DRR] A TIME INTERVAL *is a temporal quale of* a PERDURANT, an ENDURANT or a QUALITY. [**Dd35**; EP/NSMC] t *is a temporal quale of* x iff t *is a temporal quale of the endurant* x or t *is a temporal quale of the perdurant* x or t *is a temporal quale of the quality* x. [EP/IVL] *Is the temporal quale of* mutually implies *has for temporal quale*.

Is a temporal quale of an endurant, $ql_{T,ED}$
**Properties**
  [**Dd31**; EP/NSMC] t *is a temporal quale of an endurant* x iff x is an ENDURANT and t is the sum of the TIME INTERVAL t' such that x *participates in* a PERDURANT y *during* t'.

Is a temporal quale of a perdurant, $ql_{T,PD}$
**Properties**
  [**Dd30**; EP/NSMC] t *is a temporal quale of a perdurant* x iff x is a PERDURANT and there exists at least one z such that z is a TEMPORAL LOCATION and z *is a quality of* x and t *is the quale of* z.

Is a temporal quale of a physical quality or an abstract quality, $ql_{T,PQ \vee AQ}$
**Properties**
  [**Dd33**; EP/NSMC] t *is a temporal quale of a physical quality or an abstract quality* x iff x is a PHYSICAL QUALITY or an ABSTRACT QUALITY and there exists at least one z such that x *is a quality of* z and t *is a temporal quale of the endurant* z.

Is a temporal quale of a quality, $ql_{T,Q}$
**Properties**
  [**Dd34**; EP/NSMC] t *is a temporal quale of a quality* x iff t *is a temporal quale of the temporal quality* x or t *is a temporal quale of the physical quality or the abstract quality* x.

Is a temporal quale of a temporal quality, $ql_{T,TQ}$
**Properties**
  [**Dd32**; EP/NSMC] t *is a temporal quale of a temporal quality* x iff x is a TEMPORAL QUALITY and there exists a least one z such that x *is a quality of* z and t *is a temporal quale of the perdurant* z.

Is atomic during, **At**
**Properties**
  [EP/DR & RR] An ENDURANT *is atomic during* a TIME INTERVAL. [**Dd22**; EP/NSMC] x is *atomic during* t iff there does not exist y such that y *is a proper part of* x *during* t.

Is constantly specifically constituted by, **SK**
**Properties**
  [**Dd96**; EP/NSMC] x *is constantly specifically constituted by* y iff necessarily x *is present at* at least one t and y *constitutes* x *during* each t such that x *is present at* t.

Is present at, **PR**
**Properties**
  [EP/DDR & RR] An ENDURANT, a PERDURANT or a QUALITY *is present at* a TIME INTERVAL. [**Dd40**; EP/NSMC] x *is present at* t iff at least one t' exists such that t' *is the temporal quale of* x and t *is a part of* t'. [**Td17**; EP/NMC] x *is present at* t implies that x *is present at* every t' such that t' *is a part of* t.

Is spatio-temporally included in, $\subseteq_{ST}$
**Properties**

[**Dd46**; EP/NSMC] x *is spatio-temporally included in* y iff there exists at least one t such that x *is present at* t and x *is spatially included in* y *during* each t such that x *is present at* t.

Spatio-temporally coincides with, $\approx_{ST}$
**Properties**
[**Dd50**; EP/SLD] x *spatio-temporally coincides with* y iff x *is spatio-temporally included in* y and y *is spatio-temporally included i*n x.

Is temporally included in, $\subseteq_T$
**Properties**
[**Dd42**; EP/NSMC] x *is temporally included in* y iff there exists at least one t and one t' such that t *is a temporal quale of* x and t' *is a temporal quale of* y and t *is a part of* t'.

Is temporally properly included in, $\subset_T$
**Properties**
[EP/SL] *Is temporally properly included in* implies *is temporally included in*. [**Dd43**; EP/NSMC] x *is temporally properly included in* y iff there exists at least one t and one t' such that t *is a temporal quale of* x and t' *is a temporal quale of* y and t *is a proper part of* t'.

Temporally coincides with, $\approx_T$
**Properties**
[**Dd48**; EP/SLD] x *temporally coincides with* y iff x *is temporally included* in y and y *is temporally included in* x.

Is the life of, **lf**
**Properties**
[EP/DR & RR] a PERDURANT *is the life of* an ENDURANT. [**Dd68**; EP/NSMC] x *is the life of* y iff x is the sum of the z such that z *participates totally in* y.

Is the maximal participant of, **mpc**
**Properties**
[EP/DR & RR] an ENDURANT *is the maximal participant of* a PERDURANT. [**Dd66**; EP/NSMC] x *is the maximal participant of* y iff x is the sum of the z such that z *participates totally in* y.

Is the maximal physical participant of, **mppc**
**Properties**
[EP/DR & RR] a PHYSICAL ENDURANT *is the maximal physical participant of* a PERDURANT. [**Dd67**; EP/NSMC] x *is the maximal physical participant of* y iff x is the sum of the z such that z *participates totally in* y and z is a PHYSICAL ENDURANT.

Is the quale of, **ql**
**Properties**
[**Ad52**; EP/DR & RR] A TEMPORAL REGION *is the quale of* a TEMPORAL QUALITY. [**Ad54**; EP/NMC] x *is the quale of* y implies that if x' *is the quale of* y then x is equal to x'. [EP/IVL] *Is the quale of* mutually implies *has for quale*.

Overlaps with, **O**
**Properties**
[EP/DDR & DRR] An ABSTRACT or a PERDURANT *overlaps with* an ABSTRACT or a PERDURANT. [**Dd15**; EP/NSMC] x *overlaps with* y iff at least one z exists such that z *is a part of* x and z *is a part of* y.

Participates constantly in, **PC$_C$**
**Properties**
[EP/DR & RR] An ENDURANT *participates constantly in* a PERDURANT. [**Dd63**; EP/NSMC] x *participates constantly in* y iff at least one t exists such that y *is present at* t and x *participates in* y *during* each t such that y *is present at* t.

Participates totally in, **PC$_T$**
**Properties**
[EP/DR & RR] An ENDURANT *participates totally in* a PERDURANT. [**Dd65**; EP/NSMC] x *participates totally in* y iff at least one t exists such that t *is a temporal quale of* y and x *participates totally in* y *during* t.

Temporally overlaps with, **O$_T$**
**Properties**
[**Dd52**; EP/NSMC] x *temporally overlaps with* y iff there exists at least one t and one t' such that t *is a temporal quale of* x and t' *is a temporal quale of* y and t *overlaps with* t'.

## A2.4 Ternary relations

Constitutes during, **K**
**Properties**
[**Ad20**; EP/DR1 & DR2 & R3] An ENDURANT or a PERDURANT *constitutes* an ENDURANT or a PERDURANT *during* a TIME INTERVAL. [**Ad24**; EP/NMC] x *constitutes* y *during* t implies that y does not *constitutes* x *during* t. [**Ad25**; EP/NMC] x *constitutes* y *during* t implies that if y *constitutes* z *during* that t, then x *constitutes* z *during* also that t. [**Ad26a**; EP/NMC] x *constitutes* y *during* t implies that x *is present at* that t. [**Ad26b**; EP/NMC] x *constitutes* y *during* t implies that y *is present at* that t. [**Ad27**; EP/NSMC] x *constitutes* y *during* t iff x *constitutes* y *during* every t' such that t' *is a part of* t. [**Ad29**; EP/NMC] x *constitutes* y *during* t implies that if y' *is a part of* y *during* t then there exists at least one x' such that x' *is a part of* x *during* t and x' *constitutes* y' *during* t. [EP/IVL] *Constitutes during* mutually implies *has for constituent during*.

Constitutes directly during, **DK**
**Properties**
[**Dd95**; EP/SLD] x *constitutes directly* y *during* t iff x *constitutes* y *during* t and there does not exist z such that x *constitutes* z *during* t and z *constitutes* y *during* t.

Depends spatially and specifically on during, **SDt$_S$**
**Properties**
[**Dd88**; EP/NSMC] x *depends spatially and specifically on* y *during* t iff x *depends spatially and specifically on* y and x *is present at* t.

Has for constituent during, **Kinv**
**Properties**
[EP/DR1 & DR2 & R3] An ENDURANT or a PERDURANT *has for constituent* an ENDURANT or a PERDURANT *during* a TIME INTERVAL. [EP/IVL] *Has for constituent during* mutually implies *constitutes during*.

Has for part during, **Pinv**
**Properties**
[EP/R1 & R2 & R3] An ENDURANT *has for part* an ENDURANT *during* a TIME INTERVAL. [EP/IVL] *Has for part during* mutually implies *is a part of during*.

Has for participant during, **PCinv**
**Properties**
[EP/R1 & R2 & R3] A PERDURANT *has for participant* an ENDURANT *during* a TIME INTERVAL. [EP/IVL] *Has for participant during* mutually implies *participates in during*.

Has for quale during, **qlinv**
**Properties**
[EP/DR1 & DR2 & R3] A PHYSICAL QUALITY or an ABSTRACT QUALITY *has for quale* a PHYSICAL REGION or an ABSTRACT REGION *during* a TIME INTERVAL. [EP/IVL] *Has for quale during* mutually implies *is the quale of during*.

Has for spatial quale during, **ql$_S$inv**
**Properties**
[EP/DR1 & R2 & R3] A PHYSICAL ENDURANT, a PHYSICAL QUALITY or a PERDURANT, *has for spatial quale* a SPACE REGION *during* a TIME INTERVAL. [EP/IVL] *Has for spatial quale during* mutually implies *is a spatial quale of during*.

Is a part of during, **P**
**Properties**
[**Ad10**; EP/R1 & R2 & R3] An ENDURANT *is a part of* an ENDURANT *during* a TIME INTERVAL. [**Ad13**; EP/NMC] x *is a part of* y *during* t implies that if y *is a part of* a z *during* t then x *is a part of* that z *during* t. [**Ad17a**; EP/NMC] x *is a part of* y *during* t implies that x *is present at* that t. [**Ad17b**; EP/NMC] x *is a part of* y *during* t implies that y *is present at* that t. [**Ad18**; EP/NMC] x *is a part of* y *during* t implies that for each t' such that t' *is a part of* t, x *is a part of* y *during* t'. [EP/IVL] *Is a part of during* mutually implies *has for part during*.

Coincides with during, ≡$_t$
**Properties**
[**Dd24**; EP/SLD] x *coincides with* y *during t* iff x *is a part of* y *during* t and y *is a part of* x *during* t.

Is an atomic part of during, **AtP**
**Properties**
[**Dd23**; EP/SLD] x *is an atomic part of* y *during* t iff x *is a part of* y *during* t and x is *atomic during* t.

Is a proper part of during, **PP**

**Properties**
  [**Dd20**; EP/SLD] x *is a proper part of* y *during* t iff x *is a part of* y *during* t and not y *is a part of* x *during* t.

Is a quale of during, **ql**
**Properties**
  [**Ad58**; EP/DR1 & DR2 & R3] A PHYSICAL REGION or an ABSTRACT REGION *is the quale* of a PHYSICAL QUALITY or an ABSTRACT QUALITY *during* a TIME INTERVAL. [**Ad65**; EP/NMC] x *is the quale of* y *during* t implies that y *is present at* t. [**Ad66**; EP/NSMC] x *is the quale of* y *during* t iff x *is the quale of* y *during* every t' such that t' *is a part of* t. [EP/IVL] *Is the quale during* mutually implies *has for quale during*.

Is a spatial quale of a physical endurant during, **ql$_{S,PED}$**
**Properties**
  [**Dd36**; EP/NSMC] s *is a spatial quale of a physical endurant* x *during* t iff x is a PHYSICAL ENDURANT and there exists at least one z such that z is a SPATIAL LOCATION and z *is a quality of* x and s *is a quale of* z *during* t.

Is a spatial quale of a perdurant during, **ql$_{S,PD}$**
**Properties**
  [**Dd38**; EP/NSMC] s *is a spatial quale of a perdurant* x *during* t iff x is a PERDURANT and there exists at least one z such that z *is the maximal physical participant* of x and s *is a spatial quale of the physical endurant* z *during* t.

Is a spatial quale of a physical quality during, **ql$_{S,PQ}$**
**Properties**
  [**Dd37**; EP/NSMC] s *is a spatial quale of a physical quality* x *during* t iff x is a PHYSICAL QUALITY and there exists at least one z such that x *is a quality of* z and s *is a spatial quale of the physical perdurant* z *during* t.

Is present in at, **PR**
**Properties**
  [EP/DR1 & R2 & R3] A PHYSICAL ENDURANT, a PHYSICAL QUALITY or a PERDURANT *is present in* a SPACE REGION *at* a TIME INTERVAL. [**Dd41**; EP/NSMC] x *is present in* s *at* t iff x *is present at* t and at least one s' exists such that s' *is the spatial quale of* x *during* t and s *is a part of* s'. [**Td18**; EP/NMC] if x *is present in* s *at* t then x *is present at* t.

Is spatio-temporally included in during, $\subseteq_{ST,t}$
**Properties**
  [**Dd47**; EP/NSMC] x *is spatio-temporally included in* y *during* t iff x *is present at* t and x *is spatially included in* y *during* each t' such that t' *is an atomic part of* t.

  Spatio-temporally coincides with during, $\approx_{ST,t}$
  **Properties**
    [EP/SL] x *spatio-temporally coincides with* y *during* t implies x *is spatio-temporally included in* y *during* t. [**Dd51**; EP/NSMC] x *spatio-temporally coincides with* y *during* t iff x *is present at* t and x *temporally spatially coincides with* y *during* each t' such that t' *is an atomic part of* t.

Is spatially included in during, $\subseteq_{S,t}$
**Properties**
  [**Dd44**; EP/NSMC] x *is spatially included in* y *during* t iff there exists at least one s and one s' such that s *is a spatial quale of* x *during* t and s' *is a spatial quale of* y *during* t and s *is a part of* s'.

Is spatially properly included in during, $\subset_{S,t}$
**Properties**
  [EP/SL] *Is spatially properly included in during* implies *is spatially included in during*.
  [**Dd45**; EP/NSMC] x *is spatially properly included in* y *during* t iff there exists at least one s and one s' such that s *is a spatial quale of* x *during* t and s' *is a spatial quale of* y *during* t and s *is a proper part of* s'.

Spatially coincides with during, $\approx_{S,t}$
**Properties**
  [**Dd49**; EP/SLD] x *spatially coincides with* y *during* t iff x *is spatially included* in y *during* t and y *is spatially included in* x *during* t.

Is the binary sum of, +
**Properties**
  [**Dd18**; EP/NSMC] z *is the binary sum of* x and y iff z is such that every w which *overlaps with* z either *overlaps with* x or y.

Is the binary constant sum of, $+_{te}$
**Properties**
  [**Dd26**; EP/NSMC] z *is the binary constant sum of* x and y iff z is such that every w which *overlaps with* z *during* every t either *overlaps with* x or y *during* that t.

Is a spatial quale of during, $ql_S$
**Properties**
  [EP/R1 & DR2 & R3] A SPACE REGION *is a spatial quale of* a PHYSICAL ENDURANT, a PHYSICAL QUALITY or a PERDURANT, *during* a TIME INTERVAL. [**Dd39**; EP/NSMC] s *is a spatial quale of* x *during* t iff s *is a spatial quale of the physical endurant* x *during* t or s *is a spatial quale of the physical quality* x *during* t or s *is a spatial quale of the perdurant* x during t. [EP/IVL] *Is a spatial quale of during* mutually implies *has for spatial quale during*.

Overlaps with during, **O**
**Properties**
  [EP/R1 & R2 & R3] An ENDURANT *overlaps with* an ENDURANT *during* a TIME INTERVAL. [**Dd21**; EP/NSMC] x *overlaps with* y during *t* iff at least one z exists such that z *is a part of* x *during* t and z *is a part of* y *during* t.

Participates in during, **PC**
**Properties**
  [**Ad33**; EP/R1 & R2 & R3] An ENDURANT *participates in* a PERDURANT *during* a TIME INTERVAL. [**Ad36a**; EP/NMC] x *participates in* y *during* t implies that x *is present at* that t. [**Ad36b**; EP/NMC] x *participates in* y *during* t implies that y *is present at* that t.

[**Ad37**; EP/NMC] x *participates in* y *during* t implies that x *participates in* y *during* each t' such that t' *is a part of* t. [**Td7**; EP/NMC] x *participates in* y *during* t implies that y does not *participate in* x *during* t. [EP/IVL] *Participates in during* mutually implies *has for participant during*.

**Comment**
[EX] A person, which is an ENDURANT, may participate in a discussion, which is a PERDURANT. A person's life is also a PERDURANT, in which a person participates throughout its all duration.

Participates totally in during, **PC$_T$**
**Properties**
[EP/R1 & R2 & R3] An ENDURANT *participates totally in* a PERDURANT *during* a TIME INTERVAL. [**Dd64**; EP/NSMC] x *participates totally in* y *during* t iff for every z such that z *is a part of* y and z *is present at* t, x *participates in* z *during* t.

Spatially overlaps with during, **O$_{S,t}$**
**Properties**
[**Dd53**; EP/NSMC] x *spatially overlaps with* y *during* t iff there exists at least one s and one s' such that s *is a spatial quale of* x *during* t and s' *is a spatial quale of* y *during* t and s *overlaps with* s'.

## A2.5 Meta-Concepts

Anti-Atomic, **AT˜**
**Properties**
[**Dd62**; EP/SLD] An ANTI-ATOMIC (property) is a PROPERTY that *is subsumed by* PERDURANT and all of whose instances are necessarily not *atomic*.
**Comment**
[DEF] AT˜($\phi$) =$_{def}$ SB(PD,$\phi$) $\wedge$ nec$\forall$x($\phi$(x) $\rightarrow$ ¬At(x)).

Anti-Cumulative, **CM˜**
**Properties**
[**Dd58**; EP/SLD] An ANTI-CUMULATIVE (property) is a PROPERTY that *is subsumed by* PERDURANT and which does not hold of the mereological sum of two of its instances which are not part of one another.
**Comment**
[DEF] CM˜($\phi$) =$_{def}$ SB(PD,$\phi$) $\wedge$ nec$\forall$x,y(($\phi$(x) $\wedge \phi$(y) $\wedge$ ¬P(x,y) $\wedge$ ¬P(y,x)) $\rightarrow$ ¬$\phi$(x + y)).

Anti-Homeomerous, **HOM˜**
**Properties**
[**Dd60**; EP/SLD] an ANTI-HOMEOMEROUS (property) is a PROPERTY that is subsumed by PERDURANT and that does not hold for at least one temporal part of all its instances.
**Comment**
[DEF] HOM˜($\phi$) =$_{def}$ SB(PD,$\phi$) $\wedge$ nec$\forall$x($\phi$(x) $\rightarrow \exists$y(P$_T$(y,x) $\wedge$ ¬$\phi$(y)).

Anti-Rigid, **~R**
**Properties**

[EP/SLD] An ANTI-RIGID (property) is a PROPERTY that is not essential to all its instances.
**Comment**
  [DEF] ~R(ϕ) =$_{def}$ ∀x(ϕ(x) → ¬necϕ(x)).

## Anti-Unity, ~U
**Properties**
  [EP/SLD] An ANTI-UNITY (property) is a PROPERTY all of whose instances are not wholes.

## Atomic, AT
**Properties**
  [**Dd61**; EP/SLD] An ATOMIC (property) is a PROPERTY that *is subsumed by* PERDURANT and all of whose instances are necessarily *atomic*.
**Comment**
  [DEF] AT(ϕ) =$_{def}$ SB(PD,ϕ) ∧ nec∀x(ϕ(x) → At(x )).

## Carries an identity criterion, Sortal, +I
**Properties**
  [EP/SLD] A (property) CARRYING AN IDENTITY CRITERION, or "SORTAL", is a PROPERTY for which a relation exists that allows deciding necessarily and sufficiently whether two instances of the PROPERTY are equal.
**Comment**
  [DEF] +I(ϕ) =$_{def}$ ϕ(x) ∧ ϕ(y) → (ρ(x,y) ↔ x = y).

## Carries a common unity criterion, +U
**Properties**
  [EP/SLD] A (property) CARRYING A COMMON UNITY CRITERION is a PROPERTY for which there exists a single equivalence RELATION such that each instance of the PROPERTY is a whole under the RELATION.

## Cumulative, CM
**Properties**
  [**Dd57**; EP/SLD] A CUMULATIVE (property) is a PROPERTY that *is subsumed by* PERDURANT and which holds of the mereological sum of two of its instances.
**Comment**
  [DEF] CM(ϕ) =$_{def}$ SB(PD,ϕ) ∧ nec∀x,y((ϕ(x) ∧ϕ(y)) → ϕ(x + y)).

## Externally-dependent, +D
**Properties**
  [EP/SLD] An EXTERNALLY-DEPENDENT (property) is a PROPERTY all of whose instances necessarily imply the existence of an external instance of another property.

## Homeomerous, HOM
**Properties**
  [**Dd59**; EP/SLD] an HOMEOMEROUS (property) is a PROPERTY that is subsumed by PERDURANT and that holds of all the temporal parts of its instances.
**Comment**
  [DEF] HOM(ϕ) =$_{def}$ SB(PD,ϕ) ∧ nec∀x,y((ϕ(x) ∧P$_T$(y,x)) → ϕ(y)).

## Non-externally-dependent, -D
**Properties**
 [EP/SLD] A NON-EXTERNALLY-DEPENDENT (property) is a PROPERTY that is not EXTERNALLY-DEPENDENT.

## Non-Empty, NEP
**Properties**
 [EP/SLD] A NON-EMPTY (property) is a PROPERTY that necessarily possesses instances.
**Comment**
 [**Dd2**; DEF] $NEP(\phi) =_{def} nec\exists x(\phi(x))$.

## Strongly Non-Empty, NEP$_S$
**Properties**
 [**Dd56**; EP/SLD] A STRONGLY NON-EMPTY (property) is a NON-EMPTY (property) that *is subsumed by* PERDURANT and that necessarily possesses two instances x and y such that x *is* not *part of* y and y *is* not *part of* x.
**Comment**
 [**Dd56**; DEF] $NEP_S(\phi) =_{def} SB(PD,\phi) \wedge nec\exists x,y(\phi(x) \wedge \phi(y) \wedge \neg P(x,y) \wedge \neg P(y,x))$.

## Non-Rigid, -R
**Properties**
 [EP/SLD] A NON-RIGID (property) is a PROPERTY that is not essential to some of its instances.
**Comment**
 [DEF] $\sim R(\phi) =_{def} \exists x(\phi(x) \wedge \neg nec\phi(x))$.

## Not carries a common unity criterion, -U
**Properties**
 [EP/SLD] A (property) NOT CARRYING A COMMON UNITY CRITERION is a PROPERTY for which no single equivalence RELATION exists such that each instance of the PROPERTY is a whole under the RELATION.

## Not carries an identity criterion, -I
**Properties**
 [EP/SLD] A (property) NOT CARRYING AN IDENTITY CRITERION is a PROPERTY for which no relation exists that allows deciding, necessarily and sufficiently, whether two instances of the PROPERTY are equal.

## Rigid, RG
**Properties**
 [EP/SLD] A RIGID (property) is a PROPERTY that is essential for all its instances.
**Comment**
 [**Dd1**; DEF] $RG(\phi) =_{def} nec\forall x(\phi(x) \rightarrow nec\phi(x))$.

## Supplies an identity criterion, +O
**Properties**

[EP/SLD] A (property) SUPPLYING AN INDENTITY CRITERION is a PROPERTY that is RIGID, that CARRIES AN INDENTITY CRITERION and whose identity criterion is not carried by all the PROPERTIES *subsuming* it.

## A2.6 Meta-Relations

Is a non-trivial Partition of, **PT**
**Properties**
[**Dd13**; EP/NSMC] A collection $\phi_1, \ldots \phi_n$ *is a non-trivial Partition of* $\psi$ iff none of the $\phi_i$ *is equal to* $\psi$, all $\phi_i$ and $\phi_j$ *are disjoint* and being an instance of $\psi$ necessarily amounts to being an instance of one of the $\phi_i$.

Is constituted by, **K**
**Properties**
[**Dd99**; EP/NSMC] $\varphi$ *is constituted by* $\psi$ iff $\varphi$ is *constantly specifically constituted by* $\psi$ or $\varphi$ *is constantly generically constituted by* $\psi$.

Is disjoint with, **DJ**
**Properties**
[**Dd3**; EP/NSMC] $\phi$ *is disjoint with* $\psi$ iff necessarily the two properties have no instance in common.
**Comment**
[DEF] $DJ(\phi,\psi) =_{def} nec \neg \exists x(\phi(x) \wedge \psi(x))$.

Constantly depends on, **D**
**Properties**
[EP/SL] $\phi$ *constantly depends on* $\psi$ implies that $\varphi$ *is disjoint with* $\psi$. [**Dd72**; EP/NSMC] $\phi$ *constantly depends on* $\psi$ iff $\phi$ *specifically constantly depends on* $\psi$ or $\phi$ *generically constantly depends on* $\psi$.

One-sided constantly depends on, **OD**
**Properties**
[**Dd73**; EP/SLD] $\phi$ *one-sided constantly depends on* $\psi$ iff $\phi$ *constantly depends on* $\psi$ and not $\psi$ *constantly depends on* $\phi$.

Generically constantly depends on, **GD**
**Properties**
[**Dd71**; EP/SLD] $\phi$ *generically constantly depends on* $\psi$ iff $\phi$ *is disjoint with* $\psi$ and necessarily every $\phi$er *is present at* a t and for every $\phi$er which *is present at* an atomic t there exists a $\psi$er which *is present at* that t. [**Td10**; EP/SMC] $\varphi$ *generically constantly depends on* $\psi$ and $\psi$ *generically constantly depends on* $\rho$ and $\varphi$ *is disjoint with* $\rho$ implies that $\varphi$ *generically constantly depends on* $\rho$. [**Td11**; EP/SMC] $\varphi$ *specifically constantly depends on* $\psi$ and $\psi$ *generically constantly depends on* $\rho$ and $\varphi$ *is disjoint with* $\rho$ implies that $\varphi$ *generically constantly depends on* $\rho$. [**Td12**; EP/SMC] $\varphi$ *generically constantly depends on* $\psi$ and $\psi$ *specifically constantly depends on* $\rho$ and $\varphi$ *is disjoint with* $\rho$ implies that $\varphi$ *generically constantly depends on* $\rho$.

Generically spatially depends on, **GD$_S$**
**Properties**
[**Td14**; EP/SL] ϕ *generically spatially depends on* ψ implies that ϕ *generically constantly depends on* ψ. [**Dd84**; EP/NSMC] ϕ *generically spatially depends on* ψ iff ϕ *is disjoint with* ψ and necessarily every ϕer *is present in* a s *at* a t and for every ϕer which *is present in* a s *at* an atomic t there exists a ψer which *is present in* that s *at* that t.

Mutually generically spatially depends on, **MGD$_S$**
**Properties**
[**Dd94**; EP/SLD] φ *mutually generically spatially depends on* ψ iff φ *generically spatially depends on* ψ and ψ *generically spatially depends on* φ.

Partially generically spatially depends on, **PGD$_S$**
**Properties**
[**Dd85**; EP/SLD] ϕ *partially generically spatially depends on* ψ iff ϕ *is disjoint with* ψ and necessarily every ϕer *is present in* a s *at* a t and for every ϕer which *is present in* a s *at* an *atomic* t there exists a ψer and a s' such that s' *is a proper part of* s and the ψer *is present in* s' *at* t.

Inversely partially generically spatially depends on, **P$^{-1}$GD$_S$**
**Properties**
[**Dd86**; EP/SLD] ϕ *inversely partially generically spatially depends on* ψ iff ϕ *is disjoint with* ψ and necessarily every ϕer *is present in* a s *at* a t and for every ϕer which *is present in* a s *at* an *atomic* t there exists a ψer and a s' such that s *is a proper part of* s' and the ψer *is present in* s' *at* t.

One-sided generically spatially depends on, **OGD$_S$**
**Properties**
[**Dd92**; EP/SLD] ϕ *one-sided generically spatially depends on* ψ iff ϕ *generically spatially depends on* ψ and not ψ *constantly depends on* ϕ.

Is constantly generically constituted by, **GK**
**Properties**
[**Td3**; EP/SL] ϕ *is constantly generically constituted by* ψ implies ϕ *generically constantly depends on* ψ. [**Dd98**; EP/NSMC] ϕ *is constantly generically constituted by* ψ iff ϕ *is disjoint with* ψ and necessarily every ϕer *is present at* a t and for every ϕer which *is present at* an atomic t there exists a ψer which *constitutes* the ϕer *during* t. [**Td5**; EP/SMC] φ *is constantly generically constituted by* ψ and ψ *is constantly generically constituted by* ρ and φ *is disjoint with* ρ implies that φ *is constantly generically constituted by* ρ.

Is mutually generically constituted by, **MGK**
**Properties**
[**Dd103**; EP/SLD] φ *is mutually generically constituted by* ψ iff φ *is constantly generically constituted by* ψ and ψ *is constantly generically constituted by* φ.

Is one-sided constantly generically constituted by, **OGK**

**Properties**
    [**Dd101**; EP/SLD] φ *is one-sided constantly generically constituted by* ψ *iff* φ *is constantly generically constituted by* ψ *and not* ψ *is constituted by* φ.

## Mutually generically constantly depends on, MGD
**Properties**
    [**Dd77**; EP/SLD] ϕ *mutually generically constantly depends on* ψ *iff* ϕ *generically constantly depends on* ψ *and* ψ *generically constantly depends on* ϕ.

## One-sided generically constantly depends on, OGD
**Properties**
    [**Dd75**; EP/SLD] ϕ *one-sided generically constantly depends on* ψ *iff* ϕ *generically constantly depends on* ψ *and not* ψ *constantly depends on* ϕ.

# Specifically constantly depends on, SD
**Properties**
[**Dd70**; EP/SLD] ϕ *specifically constantly depends on* ψ *iff* ϕ *is disjoint with* ψ *and necessarily every* ϕer *depends constantly and specifically on* a ψer. [**Td9**; EP/SMC] φ *specifically constantly depends on* ψ *and* ψ *specifically constantly depends on* ρ *and* φ *is disjoint with* ρ implies that φ *specifically constantly depends on* ρ

## Is constantly specifically constituted by, SK
**Properties**
    [**Td2**; EP/SL] ϕ *is constantly specifically constituted by* ψ implies ϕ *specifically constantly depends on* ψ. [**Dd97**; EP/NSMC] ϕ *is constantly specifically constituted by* ψ iff ϕ *is disjoint with* ψ and necessarily every ϕer *is constantly specifically constituted by* a ψer. [**Td4**; EP/SMC] φ *is constantly specifically constituted by* ψ and ψ *is constantly specifically constituted by* ρ and φ *is disjoint with* ρ implies that φ *is constantly specifically constituted by* ρ.

## Is mutually specifically constituted by, MSK
**Properties**
    [**Dd102**; EP/SLD] φ *is mutually specifically constituted by* ψ iff φ *is constantly specifically constituted by* ψ and ψ *is constantly specifically constituted by* φ.

## Is one-sided constantly specifically constituted by, OSK
**Properties**
    [**Dd100**; EP/SLD] φ *is one-sided constantly specifically constituted by* ψ iff φ *is constantly specifically constituted by* ψ and not ψ *is constituted by* φ.

## Mutually specifically constantly depends on, MSD
**Properties**
    [**Dd76**; EP/SLD] ϕ *mutually specifically constantly depends on* ψ iff ϕ *specifically constantly depends on* ψ and ψ *specifically constantly depends on* ϕ.

## One-sided specifically constantly depends on, OSD
**Properties**

[**Dd74**; EP/SLD] ϕ *one-sided specifically constantly depends on* ψ *iff* ϕ *specifically constantly depends on* ψ *and not* ψ *constantly depends on* ϕ.

## Specifically spatially depends on, SD$_S$
**Properties**
[**Td13**; EP/SL] ϕ *specifically spatially depends on* ψ *implies* ϕ *specifically constantly depends on* ψ. [**Dd31**; EP/NSMC] ϕ *specifically spatially depends on* ψ *iff* ϕ *is disjoint with* ψ *and necessarily every* ϕer *depends spatially and specifically on* a ψer.

## Mutually specifically spatially depends on, MSD$_S$
**Properties**
[**Dd93**; EP/SLD] φ *mutually specifically spatially depends on* ψ *iff* φ *specifically spatially depends on* ψ *and* ψ *specifically spatially depends on* φ.

## One-sided specifically spatially depends on, OSD$_S$
**Properties**
[**Dd91**; EP/SLD] ϕ *one-sided specifically spatially depends on* ψ *iff* ϕ *specifically spatially depends on* ψ *and not* ψ *constantly depends on* ϕ.

# Is subsumed by
**Properties**
[EP/NSMC] ϕ *is subsumed by* ψ *iff being an instance of* ϕ *necessarily implies being an instance of* ψ. [EP/IVL] ϕ *is subsumed by* ψ *mutually implies* ψ *subsumes* ϕ.

# Subsumes, SB
**Properties**
[**Dd4**; EP/NSMC] ϕ *subsumes* ψ *iff being an instance of* ψ *necessarily implies being an instance of* ϕ. [EP/IVL] ϕ *subsumes* a ψ *mutually implies* ψ *is subsumed by* ϕ.
**Comment**
[DEF]  SB(ϕ,ψ) =$_{def}$ nec∀x(ψ(x) → ϕ(x)).

# Is equal to, EQ
**Properties**
[**Dd5**; EP/SLD] ϕ *is equal to* ψ *iff* ϕ *subsumes* ψ *and* ψ *subsumes* ϕ.

# Properly subsumes, PSB
**Properties**
[**Dd6**; EP/NSMC] ϕ *properly subsumes* ψ *iff* ϕ *subsumes* ψ *and* ψ *does not subsume* ϕ.